\pdfoutput=1
\documentclass[journal]{IEEEtran}

\usepackage{hyperref}       % hyperlinks
\usepackage{amsfonts}       % blackboard math symbols
\usepackage{nicefrac}       % compact symbols for 1/2, etc.
\usepackage{microtype}      % microtypography
\usepackage{xcolor}         %color text
\usepackage{graphicx}       %Handles Images
\usepackage{amsmath}
\usepackage{adjustbox}      % Size tables correctly
\usepackage{multirow}       % merges multiple cells row wise
\usepackage{caption}
\usepackage{subcaption}

\usepackage{mathtools}

\DeclarePairedDelimiter\floor{\lfloor}{\rfloor}

\begin{document}
%\setcitestyle{numbers}
\urlstyle{tt}

\title{TREND: Transferability based Robust ENsemble Design
}
\author{Deepak Ravikumar*,   Sangamesh Kodge*, Isha Garg*\\ and Kaushik~Roy,~\IEEEmembership{Fellow,~IEEE}% <-this % stops a space
\\ (* Equal contributors )
\thanks{
The authors are with the School of Electrical and Computer
Engineering, Purdue University, West Lafayette, IN 47906 USA 
(e-mail: \{dravikum, skodge, gargi, kaushik\}@purdue.edu). 
%All the codes used in this work can be found on \url{https://github.com/purdue-nrl/TREND}.
}% <-this % stops a space
}
\maketitle

\begin{abstract}
    Deep Learning models hold state-of-the-art performance in many fields, but their vulnerability to adversarial examples poses threat to their ubiquitous deployment in practical settings. Additionally, adversarial inputs generated on one classifier have been shown to transfer to other classifiers trained on similar data, which makes the attacks possible even if model parameters are not revealed to the adversary. This property of transferability has not yet been systematically studied, leading to a gap in our understanding of robustness of neural networks to adversarial inputs. In this work, we study the effect of network architecture, initialization, optimizer, input, weight and activation quantization on transferability of adversarial samples. We also study the effect of different attacks on transferability. Our experiments reveal that transferability is significantly hampered by input quantization and architectural mismatch between source and target, is unaffected by initialization but the choice of optimizer turns out to be critical. We observe that transferability is architecture-dependent for both weight and activation quantized models. To quantify transferability, we use simple metric and demonstrate the utility of the metric in designing a methodology to build ensembles with improved adversarial robustness. When attacking ensembles we observe that ``gradient domination" by a single ensemble member model hampers existing attacks. To combat this we propose a new state-of-the-art ensemble attack. We compare the proposed attack with existing attack techniques to show its effectiveness. Finally, we show that an ensemble consisting of carefully chosen diverse networks achieves better adversarial robustness than would otherwise be possible with a single network.

\end{abstract}

\section*{Impact Statement}
Deep Neural Networks (DNNs) have been successful in solving a wide variety of tasks and are believed to have a potential to revolutionize human life. However, adversarial attacks expose the brittle nature of the learned solution; one can easily generate images that fool DNNs with changes imperceptible to humans. Such adversarial attacks pose a great challenge to the deployment of DNNs in safety critical applications. A few settings where adversarial attacks can inflict considerable damage on human life are: 
\begin{itemize}
     \item Detection of bots and misinformation on social media
     \item Deployment in self driving cars
     \item Deployment in medical imaging and diagnostics
\end{itemize}
In most real world scenarios, the adversary does not have access to the model parameters. Yet, the adversary is still able to successfully craft attacks due to the transferable nature of these attacks. In this work, we systematically study transferability of these adversarial samples across models and preprocessing techniques, and propose a design methodology to build ensembles of DNNs that perform better under adversarial attacks. Our methodology offers an alternative way of building more robust model for real-world deployment.

\section{Introduction}

Deep learning has become state-of-the-art for many machine learning tasks over the past few years. Deep neural networks (DNNs) have achieved human level performance in image recognition \cite{Kriz2012, LeCun2015}. They have also been used for speech recognition \cite{deepSpeech2012} and natural language processing \cite{Andor_2016}. However, recent research \cite{szegedy2013intriguing, goodfellow2014explaining, kurakin2016adversarial} has shown the existence of small perturbations which when added to the input can cause DNNs to misclassify the input.
These adversarial perturbations are imperceptible to the human eye and are crafted with the specific intent of fooling deep neural networks into misclassifying the images. The existence of adversarial inputs has led to a considerable knowledge gap in the explainability of deep neural nets, which limits their use in safety critical applications such as malware detection \cite{largeScaleMalware} or autonomous driving systems \cite{papernot2017practical}.

The generation of adversarial inputs usually requires access to the target network (the network to be attacked) in the form of network weights or logits. However, in most practical scenarios, the adversary does not have access to the internal parameters, but is able to observe the output of the network for a given input \cite{bhagoji2018practical}. 
In such cases, previous works \cite{guo2019subspace,chen2017zoo, shiva2017simple} have shown that the adversary can train a substitute network by generating synthetic data using the outputs or the logits of the target network.  
Such query based attacks, often labeled as black-box attacks, are successful because of the transferability of adversarial perturbations: attacks crafted to fool one network often fool another network trained on similar data.

In this paper, with the motive of understanding the transferability of adversarial images between networks, we systematically study how transferability is affected by choice of network architecture, network initialization, optimizer, input, weight and activation quantization made by the defender and the attack methodology of the adversary. To account for these factors, we use a metric of transferability defined as the ratio of transferred images to the generated images. Further we propose a new state-of-the-art ensemble attack observing that ``gradient domination" by a single ensemble member model hampers existing attacks. Additionally, we show that understanding transferability aids in building robust ensembles of DNNs.

The key contributions of this work are summarised as follows:

\begin{itemize}
    \item We explore how the choice of initialization, model architecture, optimizer, quantization of input, weight and activation by the defender and attack methodology by the adversary affect the transferability of adversarial images from one DNN to another. We also make empirical suggestions on the most effective model to use, both from the point of view of an adversary and defender.
    These experiments were performed on various small and large datasets, attacked using PGD \cite{aleks2017deep}, Carlini Wagner $L_{2}$ attack \cite{carlini2016evaluating} and DeepFool \cite{moosavidezfooli2015deepfool}. 
    %Additionally, we study the effect of attack strength ($\epsilon$) on transferability; 
    
    % \item We fit the number of images that transfer from one model to another as the Cumulative Distribution Function (CDF) of an exponential distribution. Using this, we propose a simple metric of transferability as a function of attack strength ($\epsilon$), that can be estimated with as few as 4 empirical datapoints and generalizes to a continuum of epsilons;
    
    \item We devise a new state-of-the-art method for adversarially attacking an ensemble of DNNs. We identify the limitations of existing techniques and show that the proposed Average Gradient Direction (A-GD) attack achieves SoTA attack performance by comparing it with existing attack techniques.
    
    %\item We outline a methodology, TREND (Transferability based Robust ENsemble Design), which uses the proposed transferability metric to build an ensemble with higher robustness than would otherwise be possible with a single DNN.
    \item We outline a methodology, TREND (Transferability based Robust ENsemble Design), which uses set of diverse models having low transferability to build an ensemble with higher robustness than would otherwise be possible with a single DNN.
\end{itemize}

\section{Related Work}
\textbf{Attack Methodologies.} Many methods have been proposed for generating adversarial inputs. 
Some of the popular attacks in literature are Fast Gradient Sign Method (FGSM) \cite{goodfellow2014explaining}, a single step attack; Basic Iterative Method (BIM) \cite{kurakin2016adversarial}, a multi-step iterative attack; Carlini Wagner Attack (CW) \cite{carlini2016evaluating}, an optimization based attack; and Projected Gradient Descent (PGD)  \cite{aleks2017deep}, an iterative version of FGSM with a random start point within an \(\epsilon\) bound around the clean image.
There are numerous other attacks like Jacobian-based Saliency Map Attack (JSMA) \cite{papernot2015limitations}, DeepFool \cite{moosavidezfooli2015deepfool}, and Elastic-Net Attacks (EADAttack) \cite{chen2017ead}. In this work we study transferability under PGD \cite{aleks2017deep}, Carlini Wagner $L_{2}$ \cite{carlini2016evaluating} attack and DeepFool \cite{moosavidezfooli2015deepfool}. These attacks were chosen to encompass iterative, optimization and decision boundary attacks respectively and have been shown \cite{tramer2020adaptive, athalye2018obfuscated} to circumvent a diverse array of defenses. We utilize Back-Propagation through Differential Approximation (BPDA) \cite{athalye2018obfuscated} style gradient back-propagation to allow gradients to propagate through any non-differentiable functions such as input or activation quantization, wherever needed.

\textbf{Transferability.} The ability of adversarial perturbations generated on one model to successfully fool other models was first observed in \cite{szegedy2013intriguing} and subsequently in \cite{goodfellow2014explaining}. It has been  shown that  transferability is not unique to deep learning, but exists across various machine learning classifiers \cite{Papernot2016TransferabilityIM}. The authors of \cite{Papernot2016TransferabilityIM} show transferability across DNNs, k-Nearest Neighbors, Decision Trees, Support Vector Machines and Logistic Regression.
Such transfer attacks can be implemented successfully in both targeted or non-targeted scenarios \cite{Liu2017DelvingIT}. Moreover, the transferability of adversarial examples is hypothesized to  occur due to the alignment of the decision boundaries across various models  \cite{Liu2017DelvingIT}. The adversarial examples  are shown to span a contiguous subspace of high dimensionality ($\approx$25) \cite{tramer2017space} . The authors find that a significant factor of this subspace is shared between two models, which they attribute to the closeness of the decision boundaries learnt by the models.
However, authors of \cite{wu2018understanding} argue that the transfer is asymmetric and hence, cannot be explained completely by closely aligned decision boundaries.

\textbf{Ensembles.} Ensemble methods leverage the averaging effects of multiple models to make a final prediction. Model averaging methods include bagging \cite{Breiman1996} and boosting \cite{Freund1, Freund2}. We consider one of the simplest approaches to make decisions: the majority voting strategy. Ensembles are used as a tool to reduce variance in classifiers \cite{ensembleBiasVaraince}. However, we utilize them to gain adversarial robustness without significantly degrading the performance on clean images.  Recent research \cite{Abbasi2017RobustnessTA,Xu_2018,Sen2020EMPIR} has focused on developing defense strategies using ensembles. However, it has been shown \cite{HeWCCS17} that ensembles are not immune to adversarial attacks. In this paper, we take a deeper look at the link between transferability and the adversarial accuracy of ensembles, first suggested in \cite{HeWCCS17}. We show that careful selection of individual DNNs that make up the ensemble can improve the overall robustness of the ensemble.

\section{Experimental Setup} \label{sec:setup}

In this section, we perform experiments to study how network initialization, network architecture, input, weight and activation quantization affect transferability.
We study the effect of these factors independently on CIFAR-10, CIFAR-100 \cite{cifar} and ImageNet \cite{imagenet} datasets. All the models in the paper were trained using either the stochastic gradient descent (SGD) or Adam optimizer. The specific optimizer used is delineated in each case. Initial learning rate was set to $10^{-2}$ and it was scaled down by a factor of 10 at 60\% and 80\% completion using a learning rate scheduler. The models trained using SGD used a momentum of 0.9 and weight decay of $5\times10^{-4}$. The models were trained for 250 epochs on the ImageNet dataset and 400 epochs on CIFAR-10 and CIFAR100 datasets.
%The 60\% completion for CIFAR-10 and CIFAR-100 corresponds to 210\textsuperscript{th} epoch and 150\textsuperscript{th} epoch for ImageNet. 
At the end of each epoch the model was evaluated on the validation set and the model weights that achieved the best validation accuracy was saved. The model weights that achieved the best validation accuracy was used to evaluate the network performance on the test set and its accuracy was reported. Table \ref{tab:train_val_test_split} shows the training, validation and test set sizes for each dataset used.

{\renewcommand{\arraystretch}{1.3}
\begin{table}[hbt!]
\centering
\caption{Training, Validation and Test set sizes for the datasets used.}
\label{tab:train_val_test_split}
\small
\begin{adjustbox}{width=\columnwidth}
\begin{tabular}{|c|c|c|c|}
\hline
Dataset   & Train Set Size     & Validation Set Size & Test Set Size \\ \hline
CIFAR-10  & 45,000 (90\%)      & 5,000 (10\%)        & 10,000        \\ \hline
CIFAR-100 & 45,000 (90\%)      & 5,000 (10\%)        & 10,000        \\ \hline
ImageNet  & 1,249,137 (97.5\%) & 32,029 (2.5\%)      & 50,000        \\ \hline
\end{tabular}
\end{adjustbox}
\end{table}
}

% {\renewcommand{\arraystretch}{1.25}
% \begin{table*}[]
% \small
% \centering
% \caption{Baseline model accuracies in \% on CIFAR-10 and CIFAR-100 datasets for differently seeded models.}
% \begin{tabular}{|c|c|c|c|c|c|c|}
% \hline
% Dataset & \multicolumn{3}{c|}{CIFAR-10} & \multicolumn{3}{c|}{CIFAR-100} \\ \hline
% Seed & ResNet18 & VGG11 & VGG11 BN & ResNet18 & VGG11 & VGG11 BN \\ \hline
% 1 & 93.46 & 88.89 & 90.02 & 72.91 & 58.88 & 63.65 \\ 
% 2 & 93.13 & 88.72 & 89.75 & 71.82 & 59.36 & 63.23 \\ 
% 3 & 93.46 & 88.34 & 90.05 & 72.02 & 58.38 & 63.91 \\ 
% 4 & 93.36 & 88.69 & 89.81 & 72.38 & 60.19 & 63.51 \\ 
% 5 & 93.45 & 88.73 & 89.94 & 72.53 & 58.66 & 63.57 \\ 
% 6 & 93.30 & 88.59 & 89.99 & 72.65 & 58.26 & 64.05 \\ 
% 7 & 93.33 & 88.71 & 89.72 & 73.27 & 59.15 & 63.99 \\ 
% 8 & 93.46 & 88.17 & 90.19 & 72.87 & 58.25 & 63.39 \\ 
% 9 & 93.30 & 88.58 & 90.20 & 72.19 & 58.18 & 63.80 \\ 
% 10 & 93.06 & 88.37 & 89.69 & 72.87 & 58.19 & 63.68 \\ \hline 
% Mean & 93.33 $\pm$ 0.13 & 88.58 $\pm$ 0.21 & 89.94 $\pm$ 0.18 & 72.55 $\pm$ 0.43 & 58.75 $\pm$ 0.62 & 63.68 $\pm$ 0.25 \\ \hline
% \end{tabular}
% \label{tab:seed_base}
% \end{table*}
% }

{\renewcommand{\arraystretch}{1.36}
\begin{table*}[htb!]
\centering
\caption{Baseline accuracies and average number (mean $\pm$ std. dev) of adversarial images generated and transferred from source to target on CIFAR-10 (CF10) and CIFAR-100 (CF100) datasets for differently seeded models on PGD, CW $L_{2}$ and DeepFool attack.}
\begin{adjustbox}{width=\textwidth}
\begin{tabular}{|c|c|c|cccccc|cccccc|}
\hline
\multirow{2}{*}{Dataset} & \multirow{2}{*}{Arch.} & \multirow{2}{*}{Attack} & \multicolumn{6}{c|}{SGD} &  \multicolumn{6}{c|}{Adam} \\ \cline{4-15}
& & & Basline Acc. & Generated & Transferred & Trans. (\%) & $L_{2}$ & $L_{\infty}$ & Baseline Acc. &  Generated & Transferred & Trans. (\%) & $L_{2}$ & $L_{\infty}$\\ \hline
%cifar10 resnet
\multirow{9}{*}{CF10} & \multirow{4}{*}{ResNet18} 
&PGD&\multirow{4}{*}{93.33 $\pm$ 0.13} & 9085 $\pm$ 12 & 8166 $\pm$ 104  & 89.89 $\pm$ 1.14 &1.409 & 0.031 & \multirow{4}{*}{93.42 $\pm$ 0.11} & 9080 $\pm$ 12 & 8915 $\pm$ 33 & 98.19 $\pm$ 0.33& 1.430 & 0.031\\
 &  & CW ($\kappa=0$)&  & 9085 $\pm$ 12 & 349 $\pm$ 29  & 3.85 $\pm$ 0.32 & 0.178 & 0.019 & & 9084 $\pm$ 12 &284 $\pm$ 36 & 3.13 $\pm$ 0.39&   0.141 & 0.018\\
  &  & CW ($\kappa=30$)    &  & 9009 $\pm$ 19 & 7470 $\pm$ 258   & 82.91 $\pm$ 2.92 &  0.991 & 0.086 & & 9084 $\pm$ 12 & 2278 $\pm$ 316  & 25.08 $\pm$ 3.48&  0.265 & 0.033 \\
 &  & DeepFool & & 8961 $\pm$ 25 &  495 $\pm$ 54       & 5.52 $\pm$  0.61            & 0.413 & 0.008  & & 9084 $\pm$ 12 &332 $\pm$ 39 & 3.65 $\pm$ 0.43&  0.283 & 0.006\\ \cline{2-15} 
%cifar10 vgg11
 & \multirow{4}{*}{VGG11} 
& PGD & \multirow{4}{*}{88.58 $\pm$ 0.21} & 7039 $\pm$ 80&4123 $\pm$ 121 & 58.57 $\pm$ 1.65 & 1.566 & 0.031 & \multirow{4}{*}{87.67 $\pm$ 0.38} &  3288 $\pm$ 593 & 1681 $\pm$ 185 & 51.84 $\pm$ 4.28&  1.454 & 0.031\\
 &  & CW ($\kappa=0$)  &    &  8477 $\pm$ 20  &  536 $\pm$ 38 &               6.32 $\pm$ 0.45  & 0.303 & 0.032    &               &  8277 $\pm$ 37 & 161 $\pm$ 20 & 1.94 $\pm$ 0.24 & 0.232 & 0.027\\
  &  & CW ($\kappa=30$)    &  & 8055 $\pm$ 87 & 8473 $\pm$ 21  & 95.06 $\pm$ 0.95 & 1.185 & 0.129   &  & 8276 $\pm$ 37 & 491 $\pm$ 47  & 5.93 $\pm$ 0.56&   0.369 & 0.044\\
 &  & DeepFool &                            & 8117 $\pm$ 23  & 1279 $\pm$ 89  &     15.75  $\pm$  1.10   & 0.693 & 0.013                      &&8202 $\pm$ 31 &178 $\pm$ 23&2.16 $\pm$ 0.28&  0.340 & 0.009\\ \cline{2-15} 
%cifar10 vgg11bn
 & \multirow{4}{*}{VGG11BN} 
    & PGD      & \multirow{4}{*}{89.94 $\pm$ 0.18} &  8623 $\pm$ 22 & 7161 $\pm$ 97 & 83.04 $\pm$ 1.08 & 1.508 & 0.031  & \multirow{4}{*}{87.24 $\pm$ 0.17} &  8287 $\pm$ 22 & 4068 $\pm$ 189 & 49.08 $\pm$ 2.26&  1.467 & 0.031\\
 &  & CW ($\kappa=0$)    &   & 8632 $\pm$ 19 & 289 $\pm$ 19 & 3.35 $\pm$ 0.23 & 0.240 & 0.024   &  & 8288 $\pm$ 22  & 119 $\pm$ 14 & 1.42 $\pm$ 0.17&  0.201 & 0.018\\  
 &  & CW ($\kappa=30$)    &  & 8623 $\pm$ 19 &  7486 $\pm$ 111   & 86.82 $\pm$ 1.33  & 1.044 & 0.104  &  & 8288 $\pm$ 22 & 707 $\pm$ 70 & 8.53 $\pm$ 0.85 &  0.462 & 0.045\\
 &  & DeepFool & &  8576 $\pm$ 21 & 458 $\pm$ 25  & 5.34 $\pm$ 0.29 & 0.052 & 0.010  &   &8285 $\pm$ 22 & 128 $\pm$ 16&1.54 $\pm$ 0.19 &  0.323 & 0.006\\ \hline
%cifar100 resnet
\multirow{9}{*}{CF100} & \multirow{4}{*}{ResNet18} 
    & PGD & \multirow{4}{*}{72.55 $\pm$ 0.55} &  6549 $\pm$ 39  &  5794 $\pm$ 114 & 88.47 $\pm$ 1.53 & 1.394 & 0.031  & \multirow{4}{*}{71.25 $\pm$ 0.17} &  6368 $\pm$ 18 & 5913 $\pm$ 61   & 92.86 $\pm$ 0.88&  1.414 & 0.031\\ 
 &  & CW  ($\kappa=0$) &  & 6549 $\pm$ 39 &  296 $\pm$ 24 & 4.52 $\pm$ 0.36 & 0.134 & 0.012  &    & 6371 $\pm$ 18 & 335 $\pm$ 37 & 5.26 $\pm$ 0.58&  0.113 & 0.013\\  
&  & CW ($\kappa=30$)    &  & 6254 $\pm$ 65 & 5175 $\pm$ 204 & 82.76 $\pm$ 3.23 & 0.956 & 0.087  &  & 6371 $\pm$ 18 & 2303 $\pm$ 210 & 36.15 $\pm$ 3.29&  0.295 & 0.034  \\
 &  & DeepFool &  & 6542 $\pm$ 39  & 279 $\pm$ 24& 4.26  $\pm$ 0.35& 0.208 & 0.004  & & 6365 $\pm$ 18 & 343 $\pm$   34 & 5.39 $\pm$ 0.54 &  0.189 & 0.004\\ \cline{2-15} 
%cifar100 vgg11
 & \multirow{4}{*}{VGG11} 
    & PGD & \multirow{4}{*}{58.75 $\pm$ 0.62} &  4489 $\pm$ 54 & 2982 $\pm$ 60 & 66.44 $\pm$ 1.46 & 1.515 & 0.031  & \multirow{4}{*}{56.95 $\pm$ 0.43} & 2767 $\pm$ 154  & 1593 $\pm$ 79  & 57.71 $\pm$ 3.34 &  1.391 & 0.031\\  
 &  & CW  ($\kappa=0$)  &  & 4995 $\pm$ 38 & 346 $\pm$ 24 & 6.93 $\pm$ 0.51 & 0.230 & 0.023  & & 4645 $\pm$ 33 & 98 $\pm$ 12    & 2.13 $\pm$ 0.27 & 0.150 & 0.017\\
&  & CW ($\kappa=30$)    &  & 4911 $\pm$ 37 & 3254 $\pm$ 228     & 65.21 $\pm$ 4.72 & 0.872 & 0.092   &  & 4639 $\pm$ 34 & 279 $\pm$ 21  & 6.02 $\pm$ 0.45&  0.324 & 0.036 \\
 &  & DeepFool & & 4890 $\pm$ 42& 427 $\pm$ 33   & 8.73 $\pm$ 0.71 & 0.364 & 0.007   & &4598   $\pm$ 32&112 $\pm$    13&2.44 $\pm$ 0.29&  0.250 & 0.006\\ \cline{2-15} 
%cifar100 vgg11bn
 & \multirow{4}{*}{VGG11BN} 
    & PGD & \multirow{4}{*}{63.68 $\pm$ 0.25} & 5379 $\pm$ 24 & 3846 $\pm$ 65 & 71.50 $\pm$ 1.12 & 1.446 &0.031  &\multirow{4}{*}{65.27 $\pm$ 0.32} & 5187 $\pm$ 43 & 3295 $\pm$ 73  & 63.53 $\pm$ 1.26 &  1.457 & 0.031\\ 
 &  & CW  ($\kappa=0$)  &  &  5516 $\pm$ 25 & 201 $\pm$ 14 & 3.65 $\pm$ 0.24 & 0.177 & 0.016   &  &  5677 $\pm$ 30 & 203 $\pm$ 17 & 3.58 $\pm$ 0.30 & 0.173 & 0.018\\
  &  & CW ($\kappa=30$)    &  & 5487 $\pm$ 25 & 2570 $\pm$ 108 & 46.85 $\pm$ 1.93 & 0.817 & 0.081  &  & 5676 $\pm$ 30 & 863 $\pm$ 64 & 15.21 $\pm$ 1.12 &  0.396 & 0.042 \\
 &  & DeepFool & &5486 $\pm$ 24& 206 $\pm$ 14 & 3.76 $\pm$  0.25& 0.276 & 0.006  & & 5629 $\pm$ 32 &258 $\pm$ 22   & 4.58 $\pm$ 0.38 &  0.308 & 0.006\\ \hline
\end{tabular}
\end{adjustbox}
\label{tab:seed_ana}
\end{table*}
}

\textbf{Preliminaries.} Let the input to a DNN be represented by \(x\), its true class be denoted by \(t\), and the neural net classifier be represented by the function \(f(x)\). 
Then \(x' = x + \delta\), where \(-\epsilon \leq d(\delta) \leq \epsilon\), is said to be an adversarial input to the network if \(f(x) = t\) and \(f(x') \neq t\). Here, \(\epsilon\) represents the maximum allowed perturbation with respect to a distance metric  \(d(.)\), generally \(L_{\infty}\), \(L_{2}\)  or \(L_{0}\) norm. The norm is given by $||x||_{p} = (\Sigma_{i=1}^{n} |x_{i}|^{p})^{(1/p)}$ where \(p=0\) for \(L_{0}\), \(p=2\) for \(L_{2}\) and so on. Images are considered adversarial if the changes are imperceptible to humans but fool the network. To limit the allowable change, bounds are set on the perturbation norm to maintain imperceptibility. %We consider the \(L_{\infty}\)  distance between the original and the perturbed image as the measure of change.

\textbf{Transferability}. We study transferability between models by generating adversarial images on the source model and evaluating them on the target model. The number of adversarial images that transfer from the source to the target model is affected by the baseline accuracies of both the source and the target models. A source with high classification accuracy will generate more adversarial images compared to a source with low classification accuracy. A highly accurate model classifies more images correctly, therefore more images can be attacked and miss-classified. Similarly, The target model accuracy also has the same effect. More adversarial images transfer to target models with higher accuracy when compared to models with lower accuracy. 
%Second, increasing attack strength and attack iterations increase the number of adversarial images that transfer from the source to the target. 
Different datasets have different testset sizes, hence comparing transferability across different datasets also presents a challenge. 

To fairly evaluate transferability we need to use a metric that accounts for these factors. We define the transferability metric $TM$ as:
\begin{equation}
TM = \frac{f_{st}}{f_{ss}}
\label{eq:TM}
\end{equation}
Where $f_{ss}$ is the number of adversarial images generated by attacking the source model and $f_{st}$ is the number of images that transfer from the source to the target model. To account for differences in accuracies, we choose a subset of the testset that was correctly classified by both the source and the target model. The transferability metric $TM$ is a number between [0, 1] and represents a quantitative measure of the transferability between a given pair of models. The metric does not account for attack strength variation, however we found the $TM$ at a single attack strength is representative in most cases. Appendix \ref{adx:transfer_metric} fits a curve to model  transferability as a function of attack strength and shows that such a function is monotonically increasing implying $TM$ at a single attack strength is sufficient. The adversarial images  were generated using PGD for 40 iterations with an $\epsilon$ of 8/255 and step size of 0.01, Carlini Wagner $L_{2}$ attack for 100 iterations and DeepFool attack.

\textbf{Confusion Matrices (CM).} We present the results as a confusion matrix which represents the transferability metric ($TM$) obtained after performing transfer attacks between different pairs of models, see Figure \ref{fig:cifar10arch_pgd} for reference. These numbers were obtained by averaging over multiple runs across different seeds. The rows and columns of confusion matrix represent the effect of changing the target model and the source model, respectively. The deeper the color of the cells in the CM, the higher the transferability between the corresponding source and target model. The transferability metric ($TM$) can also be viewed as the factor with which the adversarial accuracy of the black box model is expected to decrease.  An adversary performing black box attacks has control over only the source model and would want to choose a model with the highest row average, in order to successfully attack a range of target models. Similarly, a defender controls only the target model and would want to choose the model with the lowest column average, in order to have the lowest transferability across any chosen source model. The averages are therefore shown alongside the corresponding rows and columns. The confusion matrices for CIFAR-10, CIFAR-100 and ImageNet are shown in Blue, Green and Red respectively.

{\renewcommand{\arraystretch}{1.5}
\begin{table*}[htb!]
\centering
\caption{Average number (mean $\pm$ std. dev) of adversarial images generated and transferred from source to target on CIFAR-10 (CF10) and CIFAR-100 (CF100) datasets for differently seeded models trained using Adam and SGD optimizer under PGD, CW $L_{2}$ and DeepFool attack.}
\begin{adjustbox}{width=\textwidth}
\begin{tabular}{|c|c|c|ccc|ccc|}
\hline
\multirow{2}{*}{Dataset} & \multirow{2}{*}{Arch.} & \multirow{2}{*}{Attack} & \multicolumn{3}{c|}{SGD as source to Adam as target} &  \multicolumn{3}{c|}{Adam as source to SGD as target} \\ \cline{4-9}
& &  & Generated & Transferred & Trans. (\%)   &  Generated & Transferred & Trans. (\%) \\ \hline
%cifar10 resnet
\multirow{6}{*}{CF10} & \multirow{2}{*}{ResNet18} 
&PGD & 9072 $\pm$ 14 & 8312 $\pm$ 69  & 91.63 $\pm$ 0.74  &9068 $\pm$ 14 & 8646 $\pm$ 76   & 95.34 $\pm$ 0.79\\
  &  & CW ($\kappa=30$)     & 8997 $\pm$ 23 & 7945 $\pm$ 133   & 88.31 $\pm$ 1.51  &   9072 $\pm$ 14 & 1349 $\pm$ 178   & 14.88 $\pm$ 1.96 \\
 \cline{2-9} 
%cifar10 vgg11
 & \multirow{2}{*}{VGG11} 
& PGD  & 7700 $\pm$ 110&5967 $\pm$ 105 & 77.50 $\pm$ 1.11 &  2829 $\pm$ 654 & 1795 $\pm$ 299 & 64.46 $\pm$ 4.28\\
  &  & CW ($\kappa=30$)      & 8340 $\pm$ 33 & 7190 $\pm$ 126  & 86.21 $\pm$ 1.46    & 8343 $\pm$ 33 & 453 $\pm$ 41  & 5.42 $\pm$ 0.48\\
  \cline{2-9} 
%cifar10 vgg11bn
 & \multirow{2}{*}{VGG11BN} 
    & PGD      &  8410 $\pm$ 24 & 5222 $\pm$ 185 & 62.09 $\pm$ 2.21   &  8417 $\pm$ 21 & 4244 $\pm$ 213 & 50.42 $\pm$ 2.49\\
 &  & CW ($\kappa=30$)      & 8408 $\pm$ 21 &  4781 $\pm$ 268   & 56.87 $\pm$ 3.20  & 8418 $\pm$ 21 & 599 $\pm$ 56 & 7.12 $\pm$ 0.67 \\
 \hline
 
%cifar100 resnet
\multirow{6}{*}{CF100} & \multirow{2}{*}{ResNet18} 
    & PGD &  6423 $\pm$ 31  &  5553 $\pm$ 121 & 86.45 $\pm$ 1.73  &  6420 $\pm$ 31 & 5742 $\pm$ 85   & 89.44 $\pm$ 1.20\\ 
 &  & CW ($\kappa=30$)      & 6130 $\pm$ 62 & 4866 $\pm$ 190 & 79.39 $\pm$ 3.08   & 6423 $\pm$ 31 & 1785 $\pm$ 152 & 27.78 $\pm$ 2.33\\
 \cline{2-9} 
%cifar100 vgg11
 & \multirow{2}{*}{VGG11} 
    & PGD  &  4284 $\pm$ 46 & 2778 $\pm$ 62 & 64.68 $\pm$ 1.61    & 2352 $\pm$ 179  & 1026 $\pm$ 53  & 43.72 $\pm$ 1.69 \\  
&  & CW ($\kappa=30$)     & 4729 $\pm$ 40 & 2018 $\pm$ 119     & 42.69 $\pm$ 2.74 & 4727 $\pm$ 40  & 231 $\pm$ 18  & 4.90 $\pm$ 0.37\\
\cline{2-9} 
%cifar100 vgg11bn
 & \multirow{2}{*}{VGG11BN} 
    & PGD  & 5423 $\pm$ 28 & 3563 $\pm$ 76 & 65.71 $\pm$ 1.30    & 5114 $\pm$ 41 & 3064 $\pm$ 83 & 59.91 $\pm$ 1.60  \\ 
  &  & CW ($\kappa=30$)    & 5529 $\pm$ 28 & 2219 $\pm$ 92 & 40.14 $\pm$ 1.61     & 5557 $\pm$ 28 & 796 $\pm$ 62 & 14.33 $\pm$ 1.10  \\
 \hline
\end{tabular}
\end{adjustbox}
\label{tab:opt_ana}
\end{table*}
}

\section{Transferability Analysis} \label{analysis}

\subsection{Attack Methodology}
%\subsubsection{Attacks}
In this work we study transferability under PGD \cite{aleks2017deep}, Carlini Wagner $L_{2}$ \cite{carlini2016evaluating} and DeepFool \cite{moosavidezfooli2015deepfool} attacks. From Table \ref{tab:seed_ana}, we observe that PGD generates adversarial images that transfer more readily compared to DeepFool or CW $L_{2}$ attack. We investigated the reason behind such poor transferability of CW $L_{2}$ ($\kappa = 0$) and DeepFool. We observed that CW and Deepfool attacks when used without a constraint on the adversarial sample confidence tend to identify low confidence examples with the perturbations tailored for a specific network and hence the transferability of such adversarial images is low. By increasing the confidence of the adversarial images generated by CW (achieved by tweaking the $\kappa$ parameter), we observed increased transferability (refer to Table \ref{tab:seed_ana}). Our observations are corroborated by the authors of the CW attack in their paper \cite{carlini2016evaluating}. However increasing transferability comes at an cost of increased perturbation distance. This is can be observed from Table \ref{tab:seed_ana} which shows the increase in $L_{2}$ and $L_{\infty}$ norms from $\kappa = 0$ and $\kappa = 30$. Increasing the transferability of DeepFool by increasing the confidence of the generated adversarial images counters the objective of the attack and hence we do not recommend DeepFool for transfer attacks and exclude it from further analysis in the paper. In the next Subsection we study the effect of optimizer and network initialization on transferability. 

\subsection{Training}
\subsubsection{Optimizer}
From Table \ref{tab:seed_ana} we observe that the choice of optimizer does have a significant effect on transferability. Table \ref{tab:seed_ana} suggests that ResNet18 networks trained with SGD seem to be less transferable compared to a ResNet18 networks trained with Adam optimizer. However the opposite is true for VGG11 and VGG11BN. Hence we are unable to make a recommendation on the choice of optimizer. Further, Table \ref{tab:opt_ana} shows the transferability between models trained using different optimizers. Interestingly, we observe an asymmetry, adversarial images generated for SGD trained models transfer well to Adam trained models however the reverse is not true. This empirical result suggests that improved ensemble diversity can be achieved by including models trained using Adam and SGD.
% We attribute this to the Adam optimizer finding very varied solutions \cite{Ashia2017}. This hypothesis is further supported by the Table \ref{tab:seed_ana} where the deviation in the transferred number of images is higher on average with Adam trained models, refer to the \% deviation column of Table \ref{tab:seed_ana}. 

\subsubsection{Network Initialization}
\label{sub:initialization}
The gradient descent algorithm is known to be sensitive to initialization \cite{kolen1991initialization}. Different parameter initializations lead training to converge to different solutions \cite{sutskever2013importance}. We investigate the effect of initialization by training ten models with different random initial seeds. To study the effect of initialization we consider ResNet18 \cite{resnet}, VGG11 and the batch normalized version of VGG11 (VGG11 BN) \cite{Simonyan2015VeryDC} architectures. 

\begin{figure*}[btp]
     \centering
     \begin{subfigure}[b]{0.46\textwidth}
         \centering
         \includegraphics[scale=0.35]{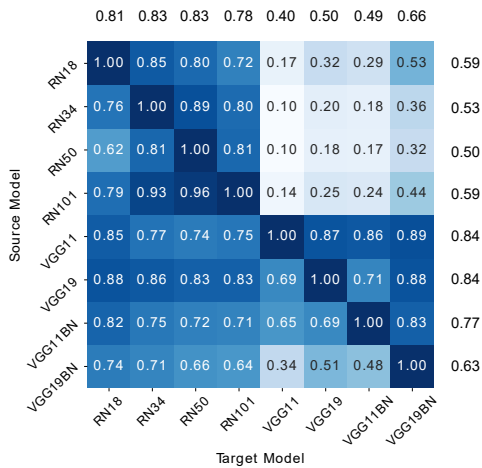}
         \caption{CIFAR-10, architecture analysis under PGD attack}
         \label{fig:cifar10arch_pgd}
         \vspace{3mm}
     \end{subfigure}
     \hfill
     \begin{subfigure}[b]{0.45\textwidth}
         \centering
         \includegraphics[scale=0.35]{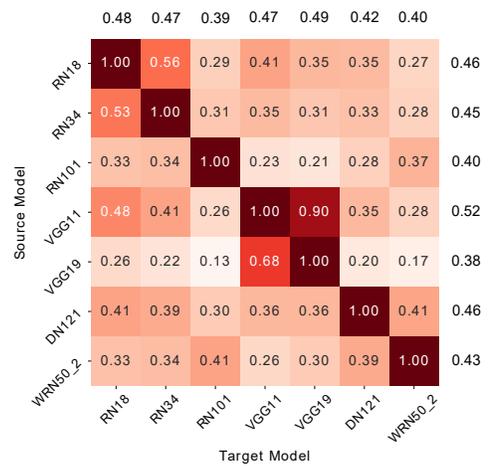}
         \caption{ImageNet, architecture analysis under PGD attack}
         \label{fig:imagenetarch_pgd}
         \vspace{3mm}
     \end{subfigure}
     \centering
     \begin{subfigure}[b]{0.46\textwidth}
         \centering
         \includegraphics[scale=0.35]{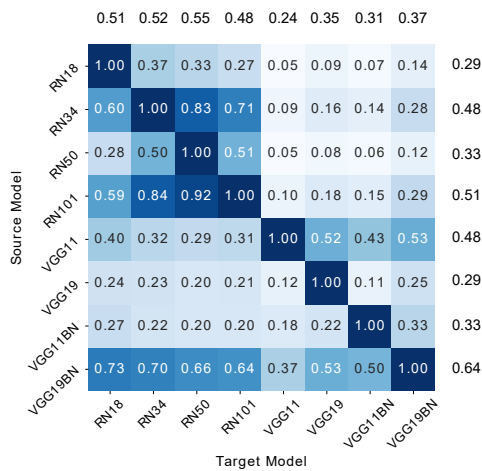}
         \caption{CIFAR-10, architecture analysis under CW $L_{2}$ attack ($\kappa = 15$)}
         \label{fig:cifar10arch_cw}
     \end{subfigure}
     \hfill
     \begin{subfigure}[b]{0.45\textwidth}
         \centering
         \includegraphics[scale=0.35]{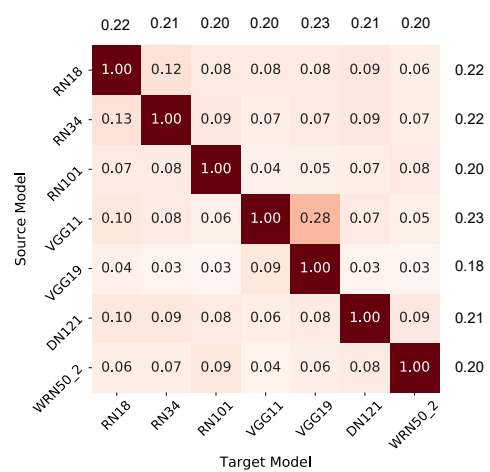}
         \caption{ImageNet, architecture analysis under CW $L_{2}$ attack ($\kappa = 30$)}
         \label{fig:imagenetarch_cw}
     \end{subfigure}
    \caption{Number of adversarial images transferred from source to target for various architectures under PGD and CW $L_{2}$ attacks.}
        \label{fig:Arch}
\end{figure*}

Table \ref{tab:seed_ana} shows the baseline accuracies and the number of adversarial images transferred from the source to the target model under PGD, CW $L_{2}$ and DeepFool attack 
%The source model for each architecture was initialized with `Seed 1'. We highlight in blue the case where the source and the target models are identical and the number of images transferred represents the number of adversarial images generated on that particular model. 
 averaged over 10 differently seeded models on CIFAR-10 and CIFAR-100. For example Table \ref{tab:seed_ana} shows ResNet18 trained using SGD on CIFAR-10 as transferring 8166 $\pm$ 104 adversarial images. This number represents the average and standard deviation over the various seeds (i.e. Seed 1 to 10) as sources and targets. Each seed was chosen as a source and the transferred number of images were used to obtain a 90 datapoint average (10 seeds, 9 targets per seed since the source seed was excluded from being the target). From Table \ref{tab:seed_ana} we observe very low deviation in the transferred adversarial images suggesting that the number of images transferred does not radically change across different seeds trained using the same optimizer. However the choice of optimizer is quite significant. \textbf{This suggests that initialization does not play a significant role in transferability but the choice of optimizer does}. Table \ref{tab:seed_ana} also seems to imply that architecture is an important consideration for transferability. We investigate this in the next subsection.

\begin{figure*}[!htb]
     \centering
     \begin{subfigure}[b]{0.48\textwidth}
         \centering
         \includegraphics[scale=0.34]{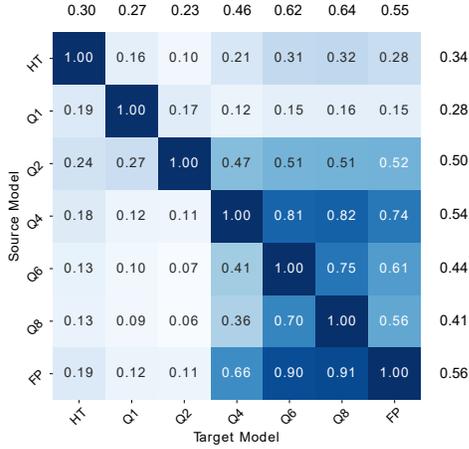}
         \caption{CIFAR-10, input quantization analysis under PGD attack}
         \label{fig:cifar10Quant_pgd}
         \vspace{3mm}
     \end{subfigure}
     \hfill
     \begin{subfigure}[b]{0.48\textwidth}
         \centering
         \includegraphics[scale=0.34]{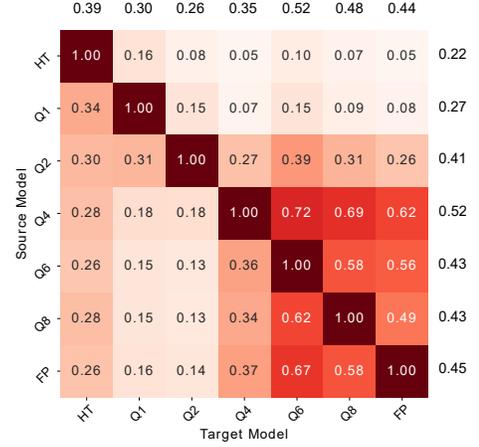}
         \caption{ImageNet, input quantization analysis under PGD attack}
         \label{fig:imagenetQuant_pgd}
         \vspace{3mm}
     \end{subfigure}
     \begin{subfigure}[b]{0.48\textwidth}
         \centering
         \includegraphics[scale=0.34]{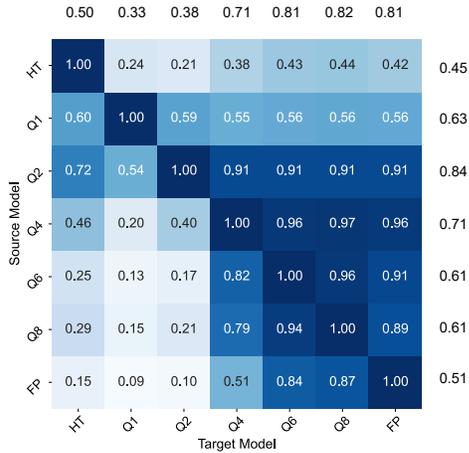}
         \caption{CIFAR-10, input quantization analysis under CW $L_{2}$ attack ($\kappa = 30$)}
         \label{fig:cifar10Quant_cw}
     \end{subfigure}
     \hfill
     \begin{subfigure}[b]{0.48\textwidth}
         \centering
         \includegraphics[scale=0.34]{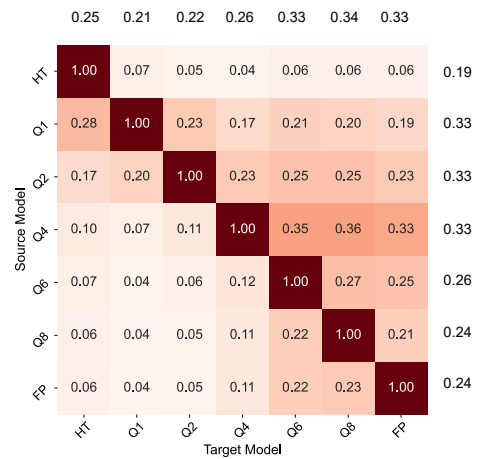}
         \caption{ImageNet, input quantization analysis under Carlini Wagner $L_{2}$ attack ($\kappa = 30$)}
         \label{fig:imagenetQuant_cw}
     \end{subfigure}
        \caption{Number of adversarial images transferred from source to target for input quantized models.}
        \label{fig:Quant}
\end{figure*}

\subsection{Architecture}
\label{sub:architecture}
We study the effect of architecture on transferability by analyzing cross model transfer of adversarial images between ResNet18 (RN18), ResNet34 (RN34), ResNet101 (RN101), VGG11, VGG19, VGG11BN, VGG19BN, DenseNet121 (DN121) \cite{densenet} and WideResNet50\_2 (WRN50\_2) \cite{wide_resnet}. 
Figure \ref{fig:Arch} shows the average (over 5 seeds) number of adversarial images transferred  from source to target under PGD attack for various architectures on CIFAR-10 and ImageNet (see Appendix \ref{adx:Arch} for CIFAR-100 and CW $L_{2}$ results).

Figure \ref{fig:Arch} shows the transferability of PGD and CW $L_{2}$ attacks across various architectures on CIFAR-10 and ImageNet dataset. Figures \ref{fig:cifar10arch_pgd} and \ref{fig:cifar10arch_cw} can be interpreted by analyzing the 4 quadrants, with each quadrant representing a family of source or target model architectures (ResNet or VGG variants). The top right quadrant of Figures \ref{fig:cifar10arch_pgd} and \ref{fig:cifar10arch_cw} is lighter than the bottom left quadrant. This implies that adversarial images generated on VGG are more transferable to ResNets than the other way around. The results for CIFAR-100 follow the same trend and are shown Appendix \ref{adx:Arch}. Surprisingly, we find that the matrices are considerably asymmetric. These findings reveal that transferability is not commutative. Another empirical observation is that the left half of Figures \ref{fig:cifar10arch_pgd} and \ref{fig:cifar10arch_cw} is the darkest, implying that ResNets are more susceptible to transfer attacks. This observation is also corroborated by \cite{Wu2020Skip} whose authors attribute this to the skip connections of ResNets and leverage this understanding to build better transfer attacks.  These trends also hold for ImageNet to certain extent. However we see a significant drop in transferability across the board when compared to CIFAR-10 and CIFAR100 this is especially true with CW $L_{2}$ attack (see Figure \ref{fig:imagenetarch_cw}). 
From Figure \ref{fig:Arch}, we observe that the column averages for ResNet18, ResNet34, VGG11 and VGG19 are among the highest. The high column average for VGG networks can be attributed to intra-family transferability, with just two numbers boosting up the column average. Hence, \textbf{VGG models are better source models for the adversary} because the row averages for VGG networks are consistently high across various datasets and provides the highest chances for successful black-box attacks. The authors of \cite{petrov2019measuring} also made similar observations with respect to VGG networks. \textbf{ResNets are easier targets and should therefore be avoided by defenders}, as they consistently show very high column averages across various datasets.

\begin{figure*}[!htb]
     \centering
     \begin{subfigure}[b]{0.48\textwidth}
         \centering
         \includegraphics[scale=0.34]{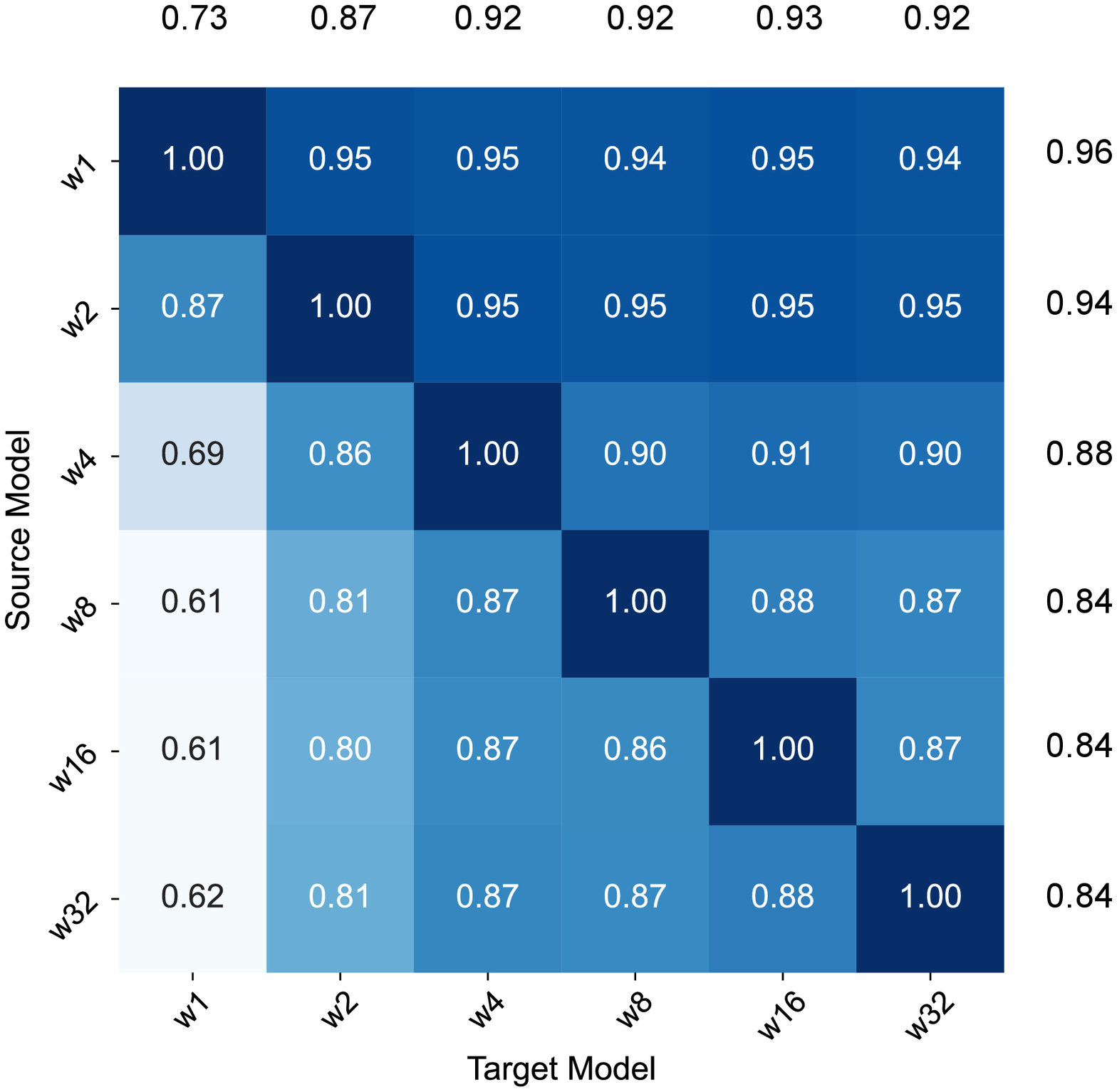}
         \caption{Weight quantized models under PGD attack}
         \label{fig:cifar10Weight_pgd}
         \vspace{3mm}
     \end{subfigure}
     \hfill
     \begin{subfigure}[b]{0.48\textwidth}
         \centering
         \includegraphics[scale=0.34]{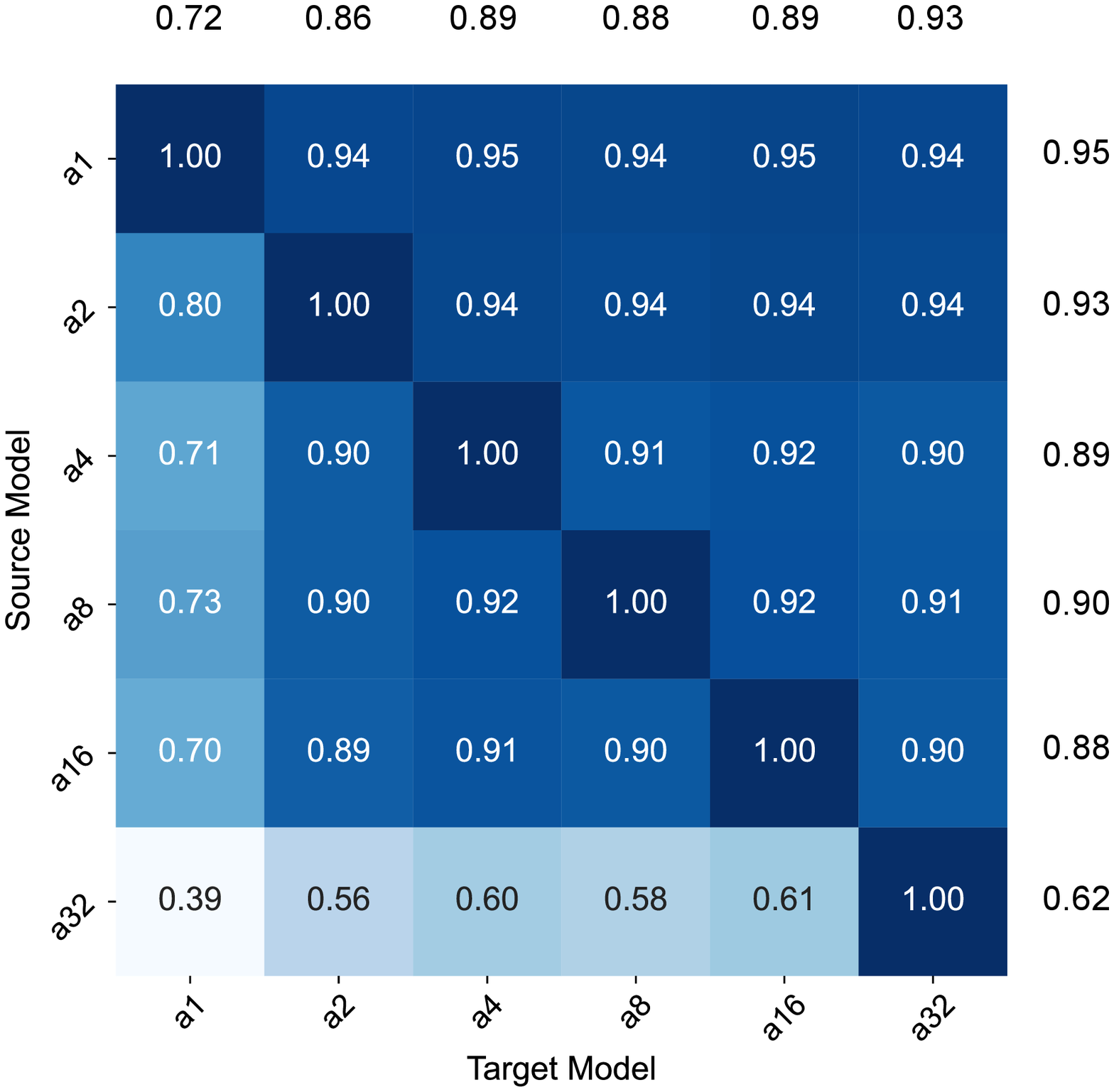}
         \caption{Activation quantized models under PGD attack}
         \label{fig:cifar10Act_pgd}
         \vspace{3mm}
     \end{subfigure}
     \begin{subfigure}[b]{0.48\textwidth}
         \centering
         \includegraphics[scale=0.34]{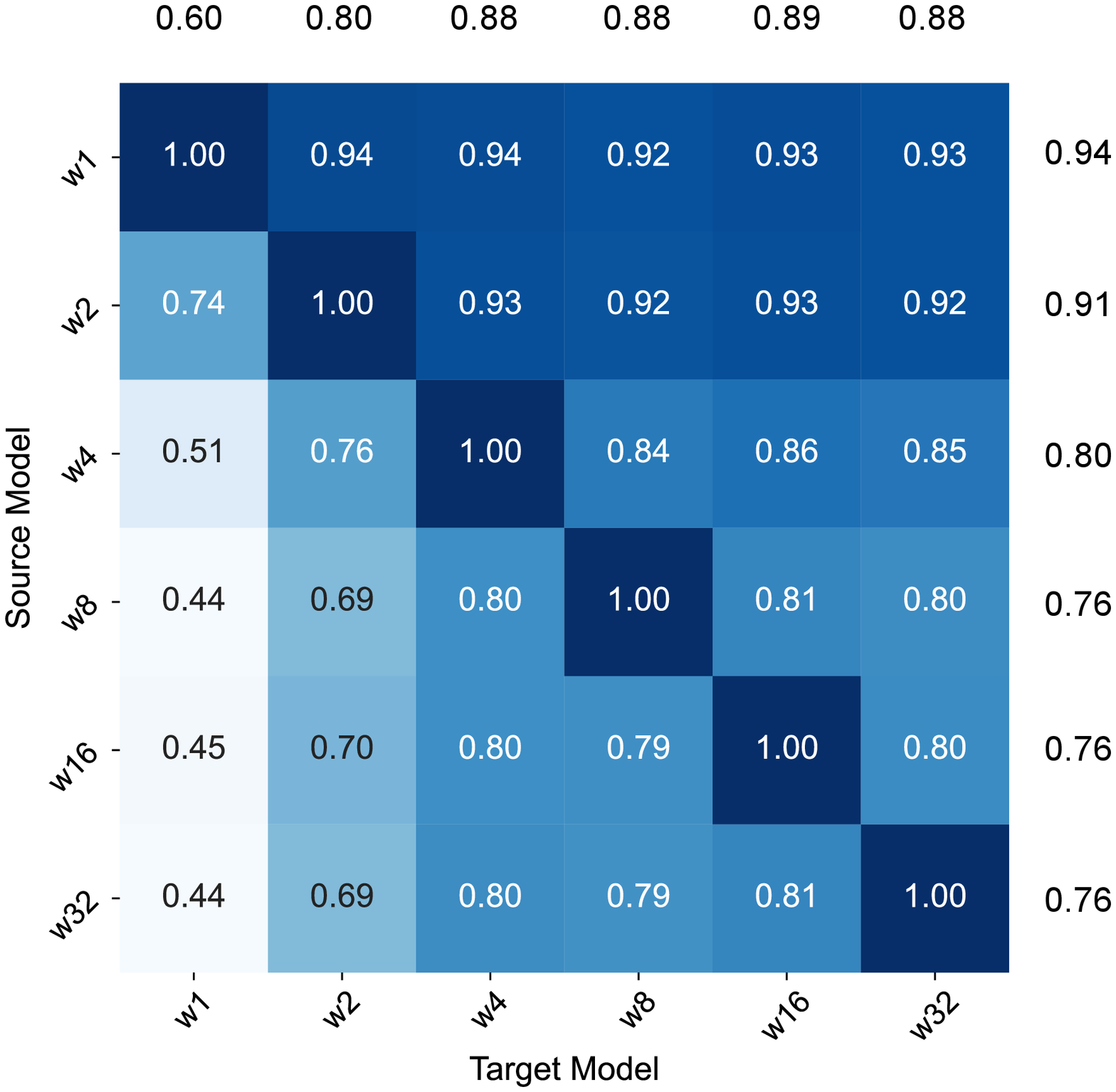}
         \caption{Weight quantized models under CW $L_{2}$ attack ($\kappa = 30$)}
         \label{fig:cifar10Weight_cw}
     \end{subfigure}
     \hfill
     \begin{subfigure}[b]{0.48\textwidth}
         \centering
         \includegraphics[scale=0.34]{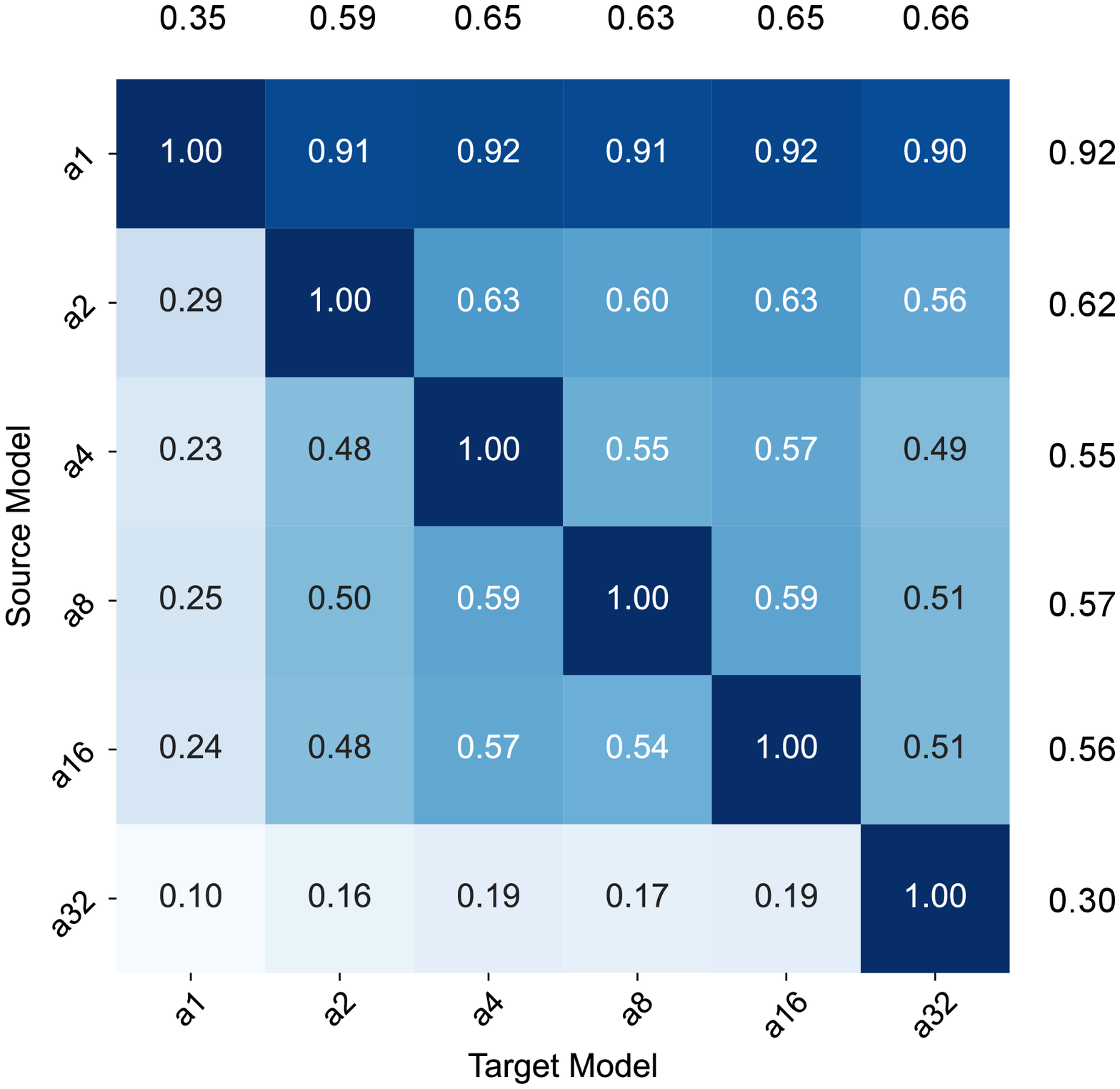}
         \caption{Activation quantized models under CW $L_{2}$ ($\kappa = 15$) attack}
         \label{fig:cifar10Act}
     \end{subfigure}
        \caption{Number of adversarial images transferred from source to target on CIFAR-10 dataset on weight and activation quantized models.}
        \label{fig:WeightandAct}
\end{figure*}

\subsection{Quantization}
\label{sub:quantization} 
Recent research \cite{Xu_2018,Sen2020EMPIR,panda2019Discretization} suggests that quantization has potential to provide robustness against adversarial images. We expect these trends to be applicable to transferability as well. Therefore, we study how input, weight and activation quantization affect transferability. 
For all experiments henceforth, our base model is ResNet18. Results for VGG11 as the base model are presented in Appendix \ref{adx:input_quant} and \ref{adx:weight_act_quant}.

\begin{figure}[!t]
      \begin{subfigure}[b]{0.3\columnwidth}
         \centering
         \includegraphics[scale=0.42]{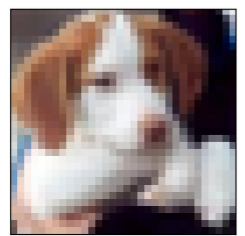}
         \caption{Image from CIFAR-10}
         \label{fig:cifar10Img}
     \end{subfigure}
     \hfill
     \begin{subfigure}[b]{0.3\columnwidth}
         \centering
         \includegraphics[scale=0.42]{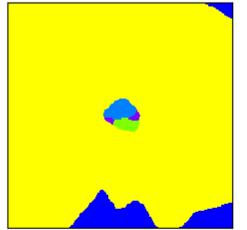}
         \caption{Decision Boundary for FP}
         \label{fig:FP_Dec_Bnd}
     \end{subfigure}
     \hfill
     \begin{subfigure}[b]{0.3\columnwidth}
         \centering
         \includegraphics[scale=0.42]{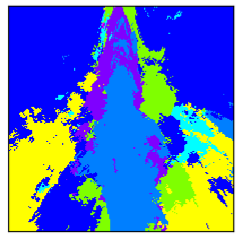}
         \caption{Decision Boundary for Q1}
         \label{fig:Q1_Dec_Bnd}
     \end{subfigure}

     \caption{Decision boundaries around image \ref{fig:cifar10Img} the of a full precision and 1-bit quantized input model.}
     \label{fig:dec_boun}
\end{figure}

\subsubsection{Input Quantization}
\label{sub:preprocessing}

Input quantization as the name suggests, quantizes the input to the network. We analyze various input bit widths ranging from 8-bits down to 1-bit per channel, per pixel for both the source and the target models. We quantize a minibatch of images using the following formula, 
\begin{equation} \label{eq:bin_size}
b = \frac{i_{max} - i_{min}}{2^{n}}
\end{equation}

\begin{equation} \label{eq:quant}
I_{quant}= \left\lfloor \frac{I-i_{min}}{b} \right\rfloor b \; + \left(\frac{1}{2}b \;+i_{min} \right)
\end{equation}

where  $I_{quant}$ is the quantized minibatch of images, $b$ is the bin width, $n$ is the bit-width used for input quantization, $i_{min}$ is the minimum and $i_{max}$ is the maximum value of the minibatch $I$. This scheme is similar to the one suggested by \cite{panda2019Discretization}, the difference being, we normalize the input before quantization. The inputs were normalized using the Z-score method. We also consider the non-linear quantization scheme of halftoning described in \cite{halftone}. The quantized models are represented by ``Q bit-width'', halftone by ``HT'' and ``FP'' refers to the full precision network. We use BPDA \cite{athalye2018obfuscated} based gradient backpropagation through the quantization scheme.
The baseline accuracies of models trained on different input bit-widths are presented in the Appendix \ref{adx:baseline_accuracy}. From Figure \ref{fig:Quant} (see Appendix \ref{adx:input_quant} for CIFAR-100 results), we see that low bit width input quantized models (HT, Q1 and Q2) have very low transferability under both PGD and CW $L_{2}$ attacks. Additionally, transferability of input quantized models is highly asymmetric. It is far easier to transfer from Q2 to various models than it is to transfer to Q2. This is also true for Q4 and HT, though to a lesser extent. To further understand the effect of input quantization, we visualize the decision boundaries of the network. The basis (i.e. x and y axes) of the visualization shown in Figure \ref{fig:dec_boun} were chosen to be the normalized adversarial gradient vector obtained from PGD (x-axis) and a random vector orthogonal to the former. Using these two vectors as the basis, the input space was traversed and the corresponding classes represented with different colors. The centers of Figures \ref{fig:FP_Dec_Bnd} and \ref{fig:Q1_Dec_Bnd} represent \ref{fig:cifar10Img} in the input space. \textbf{We observe that input quantization increases the distance to the decision boundary in most directions}, however, transfer attacks still successfully find adversarial examples.

% This leads us to conclude that \textbf{low bit width input quantization significantly reduces the success of transfer attacks}. For instance, quantizing the inputs of CIFAR-10 from FP to Q1 improves adversarial accuracy by $\sim$ 9\% between Q1 and FP. Input quantized \textbf{models with bit width greater than 2 make better source models for adversaries}. This is because these models have high row averages (see Figure \ref{fig:Quant}) which results in the highest chance for a successful black-box transfer attack. \textbf{Input quantized models with bit widths 1, 2 or HT are more robust to transfer attacks, and hence make better target models for the defenders}. This is clear from the low column averages (see Figure \ref{fig:Quant}) for these models across various datasets, provide the best chance of defense.

This leads us to conclude that \textbf{low bit width input quantization significantly reduces the success of transfer attacks}. For instance, quantizing the inputs of CIFAR-10 from FP to Q1 improves adversarial accuracy by $\sim$ 9\% between Q1 and FP. Input quantized \textbf{models with bit width greater than 2 make better source models for adversaries}. This is because these models have high row averages (see Figure \ref{fig:Quant}) which results in the highest chance for a successful black-box transfer attack. \textbf{Input quantized models with bit widths 1, 2 or HT are more robust to transfer attacks, and hence make better target models for the defenders}. This is clear from the low column averages (see Figure \ref{fig:Quant}) for these models across various datasets, provide the best chance of defense.

\subsubsection{Weight and Activation Quantization}
\label{sub:weight_quant}
In this subsection, we study how quantizing the network parameters and activations affects transferability. 
Figure \ref{fig:WeightandAct} shows the transferability among ResNet18 models with different weight and activation bit precisions for CIFAR-10 (see Appendix \ref{adx:weight_act_quant} for CIFAR-100 results).
The weight quantized models are represented by ``w bit-width'', activation quantized models by ``a bit-width'' and ``FP'' represents 32-bit full precision model.
Results for VGG weight and activation quantized models are available in Appendix \ref{adx:weight_act_quant}. We find that the trends for activation and weight quantization are highly dataset and architecture dependent. It is difficult to make generic recommendations for the adversary or the defender. We show the confusion matrices for different datasets and architectures in Appendix \ref{adx:weight_act_quant}, and the relevant ones can be consulted when making a decision.

\section{Assembling the Right Ensemble}
Recent research \cite{Abbasi2017RobustnessTA,Xu_2018,Sen2020EMPIR} suggests that ensembles provide robustness  against adversarial attacks. The robustness of the ensemble, or the lack thereof, is hypothesized to arise from the property of transferability \cite{HeWCCS17}. Since images are transferable to different models, an image meant to fool one network will fool a majority of the networks in the ensemble and therefore, the whole ensemble. We delve deeper into this hypothesis by constructing ensembles with models that have varying transferability between them, as captured by the transferability metric. We expect ensembles with models that have a high transferability metric (averaged by exchanging the source and target) between them to be less robust, as an image that fools one model will transfer well and fool the ensemble. We assemble ensembles by choosing pairs of models with low transferability metric among them, resulting in an ensemble with improved robustness. In the subsequent sections, we go into the details of how to effectively attack and design ensembles. 

\subsection{Attacking an Ensemble}
\label{sub:attack_ensemble}
\subsubsection{Existing Ensemble Attacks}
Optimization based attacks like PGD use the gradient of the loss function with respect to the input to decide the direction of change. However, the gradient for an ensemble is undefined. The most common approach is to average the gradient from each model \cite{tramer2020adaptive, HeWCCS17, strauss2017ensemble}. We refer to this as the \textbf{Direction of Average Gradient (D-AG)} attack and has been shown to be highly effective \cite{tramer2020adaptive}. Another approach is to average the gradients from the models that voted for the final predicted class as in EMPIR \cite{Sen2020EMPIR}.
 These attacks iteratively generate the adversarial images similar to PGD described by
\begin{equation}
\begin{matrix}
    x^{0} = clamp(x_{nat} + \alpha\cdot k)\\
    \\
    x^{t+1} = clamp(x^{t} + \alpha\cdot D^{t})\\

\end{matrix}
\end{equation}
where $x_{nat}$ is a natural image, $k \in \mathbb{R}^{d}$ and is sampled from $unif(-1,1)$, $d$ is the input image dimension, $x^{t}$ is the adversarial image at $t^{th}$ iteration, $\alpha$ is the $L_{\infty}$ bound for the attack, $D^{t} \in \{-1,1\}^{d}$ is the gradient direction for the ensemble at $t^{th}$ iteration and the $clamp$ function clamps its input to the image bounds $[0,1]$. The D-AG attack commonly used in literature calculates the gradient direction $D^{t}_{DAG}$ given by
\begin{equation}
\begin{matrix}
    G^{t}_{i} = \nabla_{x^{t}}L(\theta_{i}, x^{t},y)\\
    \\
    D^{t}_{DAG} = sgn(\frac{1}{N}\sum_{i=1}^{N}G^{t}_{i})
\end{matrix}
\end{equation}
where $N$ is the number of models in the ensemble, $sgn$ is the sign function and $\nabla_{x^{t}}L(\theta_{i}, x^{t},y)$ is the gradient of the $i^{th}$ member of the ensemble whose parameters are $\theta_{i}$.

\subsubsection{Proposed Attacks}
The challenge when attacking ensembles is that gradient from one of the models in the ensemble tends to dominate the average gradient, especially when the attack strength is low a phenomenon we dub ``gradient domination". This results in an adversarial direction that is unable to fool multiple models simultaneously, reducing the attack's effectiveness. Hence, we devise attacks that account for this phenomenon and show that we achieve SoTA attack success rates. Observing that the PGD attack uses the gradient direction rather than the gradient (magnitude and direction) is key to our attacks. This provides flexibility in identifying the gradient direction for the ensemble and allows our attack methods to counter the ``gradient domination" phenomenon. The first method, which we call \textbf{Unanimous Gradient Direction (U-GD)} attack chooses only those gradient directions where all the individual models' gradient directions align. The second method, \textbf{Average Gradient Direction (A-GD)}, calculates the gradient direction for the ensemble by calculating the sign of average gradient direction.

% \begin{equation}
% \begin{adjustbox}{width=\columnwidth}
% $D = \left\{ \begin{array}{cl}
% x & x = y :\space  x,\space y \in S_{i}, S_{j} \ \forall\  i \neq j; \  i,j \in [1,N] \\
% 0 & otherwise
% \end{array} \right.$
% \label{eq:signall}
% \end{adjustbox}
% \end{equation}

% \begin{equation}
% \begin{adjustbox}{width=\columnwidth}
% $sgn(G_{ensemble}) = \left\{ \begin{array}{cl}
% sgn(G_i) & sgn(G_i) = sgn(G_j) \space \forall\  i \neq j; \  \\&i,j \in [1https://www.overleaf.com/project/5e9863cc5cf8730001209a84,N] \\
% \\0 & otherwise
% \end{array} \right.$
% \label{eq:signall}
% \end{adjustbox}
% \end{equation}

\begin{figure*}[hbt!]
    \centering
    \includegraphics[scale=0.45]{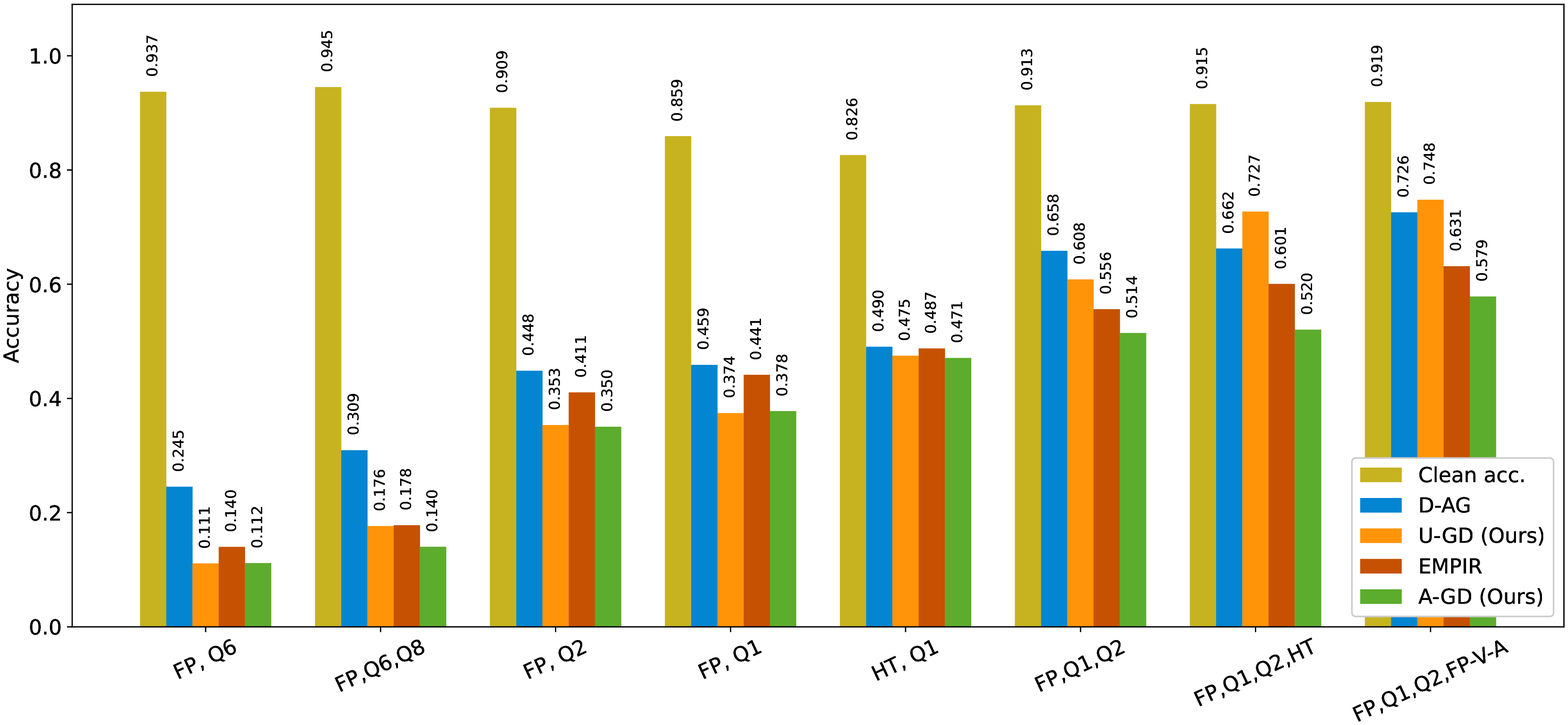}
    \caption{Accuracy of various ensembles (attack strength $\epsilon$ = 0.01) under D-AG, U-GD, A-GD and EMPIR style attack. From the plots, we see that models with low transferability form more robust ensembles. Note FP-V-A refers to a full precision VGG11 adam trained model. All other models were SGD trained ResNet18. Note across the various ensembles our A-GD attack is the most successful.}
    \label{fig:ensemble_accuracies_avg}
\end{figure*}

\textbf{Unanimous Gradient Direction (U-GD).} The U-GD attack calculates the gradient direction $D^{t}_{UGD}$ for the ensemble as given by
\begin{equation}
\begin{matrix}

S_{i}^{t} = sgn(G_{i}^{t})\\
\\
A^{t} = \frac{1}{N}(\sum_{i=1}^{N}S_{i}^{t}) \\
\\
M^{t} = \floor{|A^{t}|}{} \cdot sgn(A^{t})\\
\\
D^{t}_{UGD} = A^{t} \cdot M^{t} \\

%M = \left\{ \begin{array}{cl}
%a & : \ a \in |A| \text{ and } a =1 \\
%0 & \text{otherwise}
%\end{array} \right.
\end{matrix}
\label{eq:ugd}
\end{equation}
where $A^{t} \cdot M^{t}$ represents the element wise product of the average gradient direction $A^{t}$ with the mask $M^{t}$, $\floor{|A^{t}|}{}$ represents the floor of the absolute value of $A^{t}$, $N$ is the number of models in the ensemble and $sgn$ is the sign function.
Equation \ref{eq:ugd} translates to choosing only those gradient directions where all the individual model's gradient directions are in agreement. However, as the number of models in the ensemble increase the effectiveness of this approach decreases (refer Figures \ref{fig:ensemble_accuracies_avg} and \ref{fig:nVrob}). This is due to the strict requirement that all models must agree on the gradient direction. To overcome this we propose a simplified version of U-GD, A-GD. 

\begin{figure*}[hbt!]
    \centering
    \includegraphics[scale=0.36]{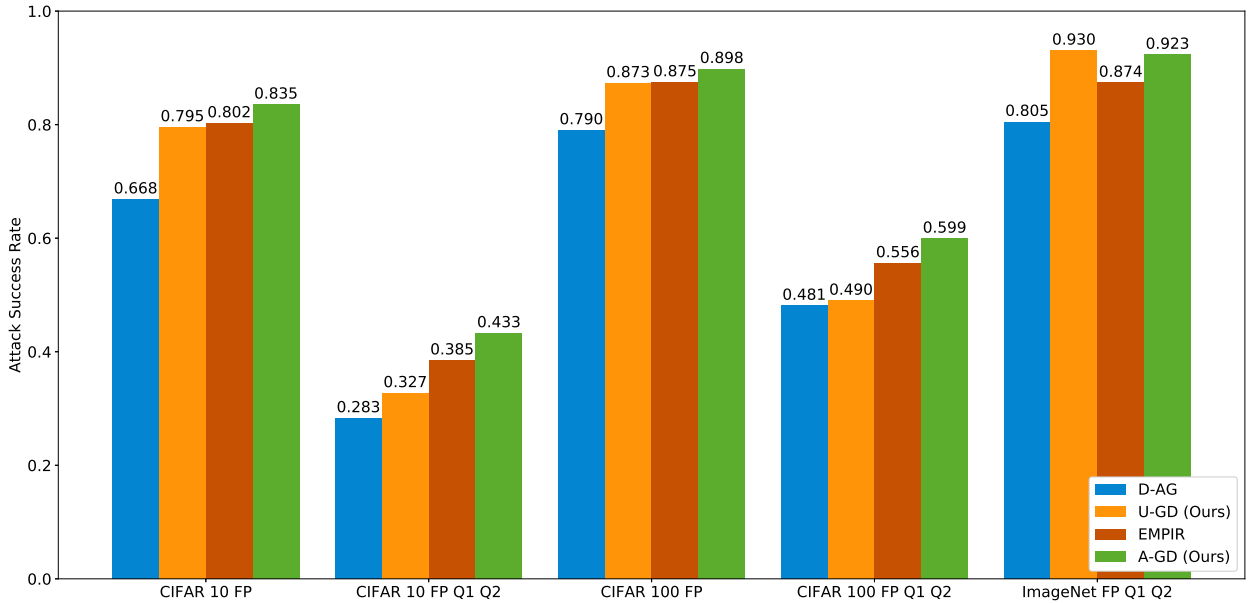}
    \caption{Attack Success Rate of different attacks at $\epsilon$ of 0.01, on CIFAR-10, CIFAR-100 and ImageNet datasets for various ensembles. Here CIFAR-10 FP and CIFAR-100 FP refers to ensembles of full precision models with different seeds.}
    \label{fig:attacks_comp}
\end{figure*}

\textbf{Average Gradient Direction (A-GD)} In this approach, the gradient direction for the ensemble is calculated by averaging the gradient direction  from each model.
\begin{equation}
D^{t}_{AGD} =  sgn(A^{t})
%sgn(G_{ensemble}) = \frac{1}{N}\sum_{m=1}^{N}sgn(G_{m})
\end{equation}
where $A^{t}$ is given in Equation \ref{eq:ugd} and $sgn$ is the sign function. We evaluate the performance of our attacks on deep ensembles which are traditionally \cite{ensembleMethods, lakshminarayanan2017simple} built with models that are trained independently and the diversity among members arises from the randomness of the initialization and of the training procedure. Figure \ref{fig:attacks_comp} compares the success rate of the proposed attack methods against D-AG \cite{tramer2020adaptive, HeWCCS17, strauss2017ensemble} and EMPIR style attack \cite{Sen2020EMPIR}.

\begin{figure}[htb!]
    \centering
    \includegraphics[scale=0.32]{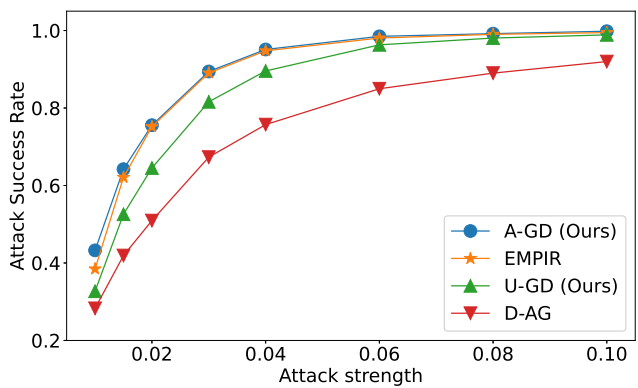}
    \caption{Attack Success Rate of different attack at various $\epsilon$ on CIFAR10 ensemble of FP, Q1, Q2.}
    \label{fig:attacks_comp_across_eps}
\end{figure}

\begin{figure}[hbt!]
    \centering
    \includegraphics[scale=0.45]{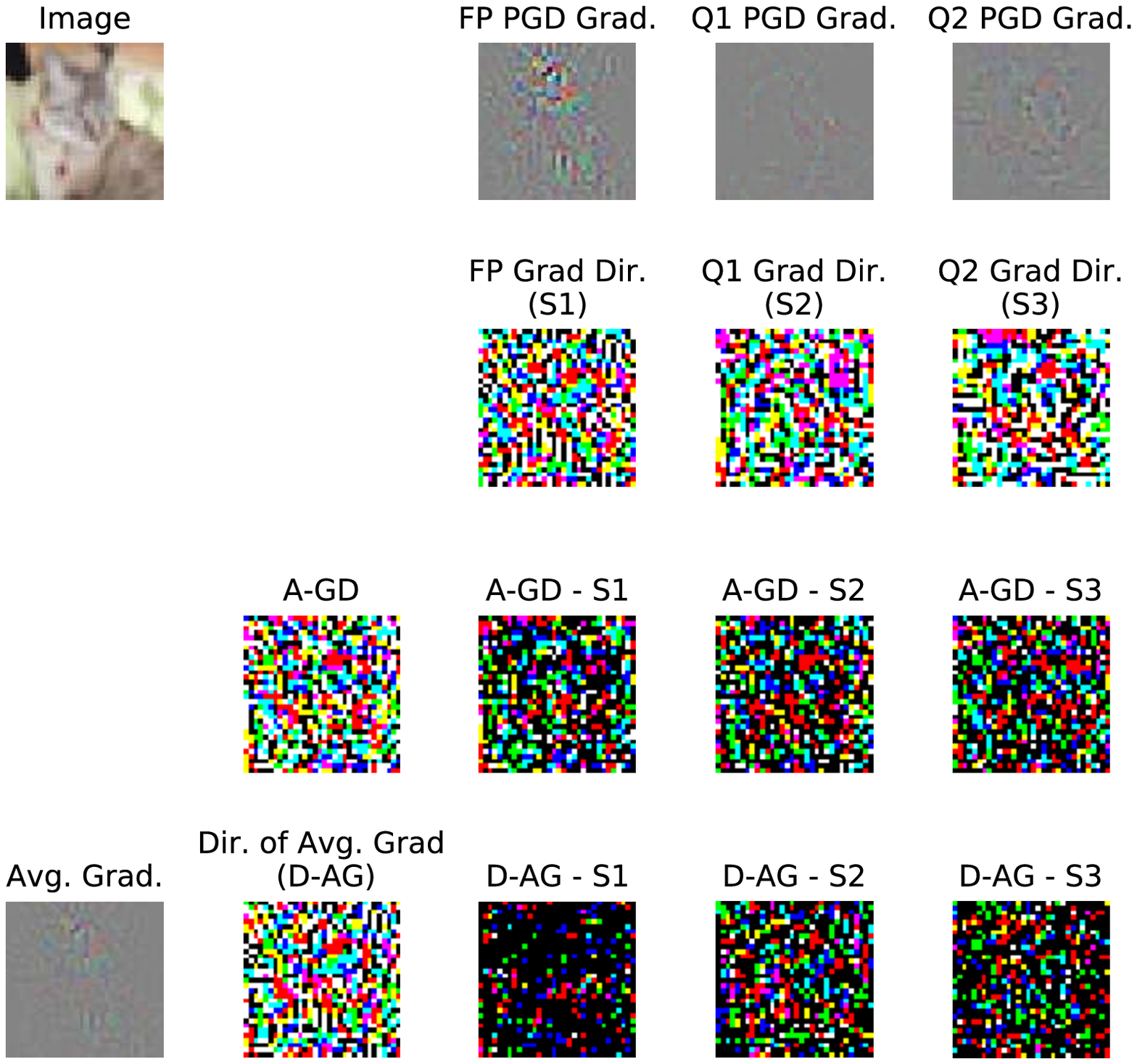}
    \caption{The direction of average gradient (D-AG) commonly used in literature is dominated by one model in the ensemble compared to our proposed method (A-GD). Gradient and gradient direction visualized as an RGB image, with black as 0 intensity and white as maximum intensity in the RGB scale.}
    \label{fig:attack_exp}
\end{figure}

Figure \ref{fig:attacks_comp_across_eps} compares the performance of various attacks across different attack strengths. We see that the proposed A-GD attack significantly outperforms other methods at lower epsilons. Other attacks get close to A-GD performance around attack strength of about 0.03 and higher. The effectiveness of the A-GD attack is because of it ability to counter the``gradient domination" problem by using the gradient direction  rather than both the magnitude and direction. This is illustrated with the help of Figure \ref{fig:attack_exp}. The first row of Figure \ref{fig:attack_exp} shows an image from CIFAR-10 and the corresponding adversarial gradients  visualized as an RGB image for FP, Q1 and Q2 ResNet18 models. The second row visualizes the individual gradient directions S1 (i.e. $S_{1}^{39}$ from Equation \ref{eq:ugd}), S2 and S3 respectively as an RGB image. The third row's first image visualizes the A-GD's gradient direction, and the rest of the row shows the difference between A-GD's direction and individual gradient direction. The final row does the same but for D-AG. The first image of the final row visualizes the average gradient from the three models, the second visualizes the gradient direction for this average. From the visualizations we can clearly see the similarity between D-AG and S1. Further we see the `D-AG - S1' is nearly all zero implying the D-AG is dominated by S1 i.e. model 1's gradient direction. Comparing this to A-GD, we observe that `A-GD - S1' is not all zero implying that the A-GD's adversarial direction is not dominated by gradient directions from a single model but is more equally shared among the member models making the A-GD attack more effective. The A-GD attack is no more compute intense than D-AG but is significantly more effective (up-to \textbf{1.56x} improvement in attack success rate). Having observed that A-GD attack achieves state-of-the-art performance, we use this attack for testing our ensembles in the subsequent sections and we recommend the use of A-DG as the benchmark for testing all future DNN ensembles.

\subsection{Robust Ensemble Design}
\begin{figure*}[htb!]
    \centering
    \includegraphics[scale=0.46]{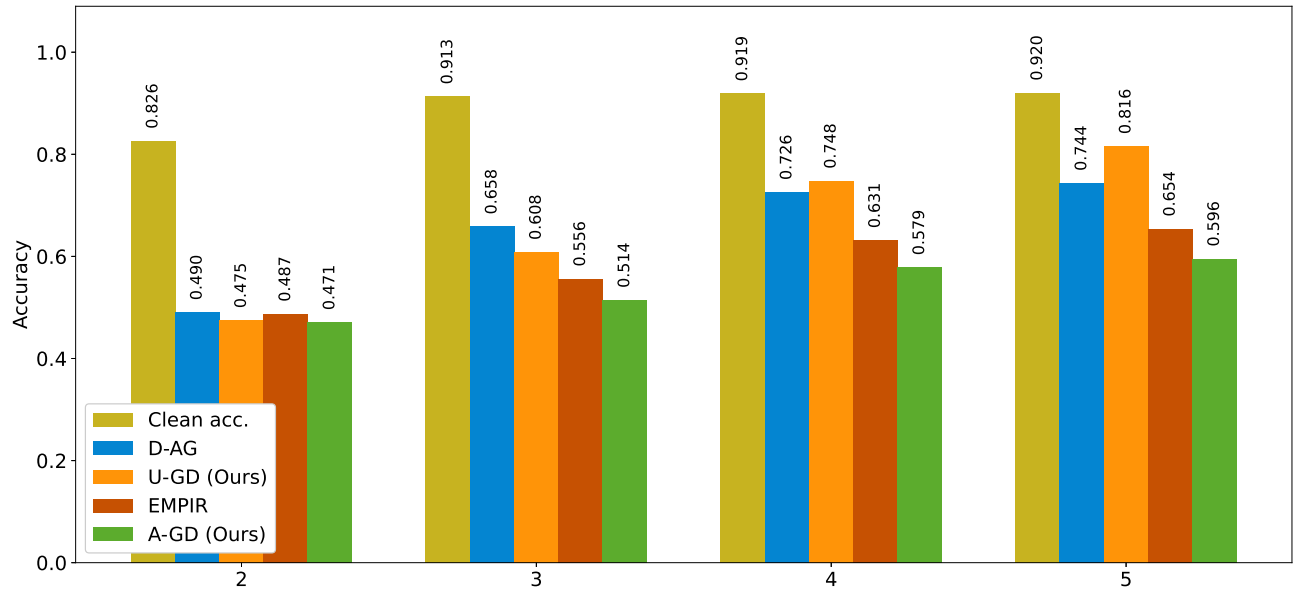}
    \caption{Number of models in a TREND built ensemble versus adversarial robustness (\% accuracy, higher is more robust) at $L_{\infty}$ bound of 0.01 on CIFAR-10 under different attacks. The robust ensembles TREND methodology choose for were 2-\{HT ResNet18, Q2 ResNet18\}, 3-\{FP ResNet18, Q1 ResNet18, Q2 ResNet18\}, 4-\{FP ResNet18, Q1 ResNet18, Q2 ResNet18, FP VGG11 adam trained\}, 5-\{FP ResNet18, Q1 ResNet18, Q2 ResNet18, FP VGG11 adam trained, FP VGG11BN adam trained\}.  }  
    \label{fig:nVrob}
\end{figure*}

In this subsection, we utilize the transferability metric introduced in Section \ref{sec:setup} to put forth a methodology TREND, to build an ensemble with improved robustness. The ensembles predicts a class using majority voting. In case of conflict (i.e. no majority vote), one of the models is chosen at random and its output is considered as the ensemble's prediction. The hypothesis that high transferability between models in an ensemble results in reduced adversarial robustness was first suggested in \cite{HeWCCS17}. We leverage this idea and build an ensemble with improved robustness by choosing models with low transferability. To identify these models, we consider a list of all pairs of individual models under consideration and calculate the transferability metric for each pair. To account for asymmetry, we average the two numbers obtained by interchanging source and target. From this list, we choose a desired number of models with the lowest transferability metric.
%Figure \ref{fig:tm_cifar10} shows the transferability metric between various input quantized configurations of ResNet18 trained on CIFAR-10. 
Ensembles built using this method were evaluated using the attacks described in Subsection \ref{sub:attack_ensemble}. Figure \ref{fig:ensemble_accuracies_avg} shows various ensembles and the corresponding accuracies. From Figure \ref{fig:ensemble_accuracies_avg} we see that the ensemble of FP-Q1-Q2, FP-Q1-Q2-HT and FP-Q1-Q2-FP-V-A consistently outperform other ensembles with respect to adversarial robustness. This trend was expected from the transferability metric for Q1, Q2 and HT input quantized models, which shows that Q1, Q2 and HT have the lowest average transferability metric. The addition of FP model boots the baseline accuracy.  The trend holds for different datasets, as shown in Appendix \ref{adx:Results_Ensemble}. Appendix \ref{adx:Results_Ensemble} also details various ensemble combinations and their adversarial accuracies under attack, and the trends expected from the transferability metric hold.
We also observe that adding more low TM models to the ensembles boosts adversarial robustness while maintaining baseline accuracy. Figure \ref{fig:nVrob} illustrates this, showing that \emph{increasing the number of models in a TREND ensemble boosts adversarial robustness under the strongest attack while maintaining baseline accuracy}. An ensemble, FP-Q1-Q2, designed using TREND is seen to be 1.30x, 1.36x and 8.43x more robust than individual Q1, Q2 and FP models correspondingly at $\epsilon$ of 0.01.
%The improved robustness is seen as an increase in $L_{\infty}$ perturbation needed to fool such ensembles when compared to ensembles consisting of models with high transferability as seen in Figure \ref{fig:ensemble_accuracies_avg}. The parameters used for PGD were the same as in Section \ref{analysis}.
% new numbers are for 40 iterations of PGD
%except the number of iterations was reduced to 20 to shorten the simulation time.

\section{Discussion and Conclusion}
 TREND is a methodology to systematically design an ensemble with improved adversarial robustness.
 In this paper we analyze the effect of DNN initialization, architecture, and input, weight and activation quantization on transferability. Our analysis suggests that initialization has no significant effect on transferability. However the choice of optimizer is critical and our experiments show that the ResNet architecture is more susceptible to transfer attacks than the other architectures considered. Quantizing the inputs significantly reduces transferability when the inputs are quantized to low bit widths (one and two bits).
Additionally, our experiments reveal that the effect of weight and activation quantization is highly dependent on the dataset.
We also observe that transferability is asymmetric. If adversarial images transfer well from source to target, the vice versa need not necessarily be true. Additionally, wherever applicable, we offer guidelines for the construction of defense and attack models based on the transferability trends observed.
%To account for the effect of attack strength, epsilon, we propose a simple transferability metric that can be estimated with very few measurements and generalizes to a continuum of epsilons. The constants that fit the distribution can be estimated with as little as 4 experimental datapoints, and capture the dependencies on all considered factors such as architecture, quantization and initialization. 
We identify that current ensemble attacks are hampered by the ``gradient domination" effect and propose two attack methods that overcome this to achieve SoTA attack performance. 
%We employ the transferability metric to design robust ensembles
%that require stronger attack strength compared to the constituting models for a similar accuracy degradation. 
Our results clarify that the adversarial robustness of an ensemble is indeed determined by how transferable an adversarial image is among the models in the ensemble. Using the transferability metric, we are able to construct ensembles with improved robustness. An TREND designed ensemble of FP-Q1-Q2 is seen to be 1.30x, 1.36x and 8.43x more robust than individual Q1, Q2 and FP models correspondingly at $\epsilon$ of 0.01.

% \section*{Broader Impact}
% DNNs have been proposed to solve many applications and are believed to have a potential to revolutionize human life. DNNs can automate many laborious and tiresome tasks enabling us to save human effort at the same time increasing the throughput of the task at hand. Due to these abilities, researchers have spent considerable time and resources to improve the performance of these networks.

% Adversarial attacks pose a great challenge to the deployment of DNNs in most safety critical applications, even with state-of-the-art defence techniques. For instance, in the following settings, adversarial attacks can inflict considerable damage on human life.
% \begin{itemize}
%     \item Detecting bots and misinformation on social media
%     \item Deployment in self driving cars
%     \item Deployment in medical imaging and diagnostics
% \end{itemize}

% Our work tackles this problem of adversarial robustness by proposing a design methodology to select the component models of an ensemble. We show that using TREND, it is possible to construct a robust ensemble. We believe our work is a positive step in achieving the reliable deployment of DNNs. 
\section*{Acknowledgement} This work was supported in part by the Center for Brain Inspired Computing (C-BRIC), one of the six centers in JUMP, a Semiconductor Research Corporation (SRC) program sponsored by DARPA, by the Semiconductor Research Corporation, the National Science Foundation, Intel Corporation, the DoD Vannevar Bush Fellowship, and by the U.S. Army Research Laboratory and the U.K. Ministry of Defence under Agreement Number W911NF-16-3-0001.

\iffalse
\bibliography{references}
% In citation order
\bibliographystyle{IEEEtran}
\fi


\newpage

\setcounter{section}{0}
\section*{Appendix}

% \section{Training Process} \label{adx:training_process}
% All the models in the paper were trained using the stochastic gradient descent optimizer with a momentum of 0.9 and weight decay of $5\times10^{-4}$. The models were trained for 250 epochs on the ImageNet dataset and 350 epochs on CIFAR-10 and CIFAR100 datasets. Initial learning rate was set to $10^{-2}$ and it was scaled down by a factor of 10 at 60\% and 80\% completion using a learning rate scheduler. The 60\% completion for CIFAR-10 and CIFAR-100 corresponds to 210\textsuperscript{th} epoch and 150\textsuperscript{th} epoch for ImageNet. At the end of each epoch the model was evaluated on the validation set and the model weights that achieved the best validation accuracy was saved. At the end of training, the model weights that achieved the best validation accuracy was used to evaluate the network performance on the test set and its accuracy was reported. Table \ref{tab:train_val_test_split} shows the training, validation and test set sizes for each dataset used.
% \begin{table}[hbt!]
% \centering
% \caption{Training, Validation and Test set sizes for the datasets used}
% \label{tab:train_val_test_split}
% \begin{tabular}{|c|c|c|c|}
% \hline
% Dataset   & Train Set Size     & Validation Set Size & Test Set Size \\ \hline
% CIFAR-10  & 45,000 (90\%)      & 5,000 (10\%)        & 10,000        \\ \hline
% CIFAR-100 & 45,000 (90\%)      & 5,000 (10\%)        & 10,000        \\ \hline
% ImageNet  & 1,249,137 (97.5\%) & 32,029 (2.5\%)      & 50,000        \\ \hline
% \end{tabular}
% \end{table}

\section{Baseline Accuracies} \label{adx:baseline_accuracy}
The baseline accuracies of the quantized models on different datasets with base architecture ResNet18 and VGG11 are shown in Table \ref{tab:quant_base_resnet} and \ref{tab:quant_base_vgg} respectively. Baselines accuries for different architecture are presented in Table \ref{tab:arch_base}

{\renewcommand{\arraystretch}{1.3}
\begin{table}[bht!]
\centering
\caption{Baseline model accuracies for SGD trained quantized inputs, weight and activations on CIFAR-10, CIFAR-100 and ImageNet with ResNet18 as baseline }
\begin{tabular}{|c|c|c|c|}
\hline
Quantization        & CIFAR-10 & CIFAR-100 & ImageNet \\ \hline
FP & 93.33 $\pm$ 0.13\%  & 72.55 $\pm$ 0.43 \%   & 55.73\%  \\ \hline
Q8 & 94.03 $\pm$ 0.14\%  & 72.38 $\pm$ 0.24\%   & 55.63\%  \\ \hline
Q6 & 93.93 $\pm$ 0.12\%  & 72.58 $\pm$ 0.26\%   & 55.39\%  \\ \hline
Q4 & 93.42 $\pm$ 0.11\%  & 71.10 $\pm$ 0.13\%   & 55.09\%  \\ \hline
Q2 & 88.22 $\pm$ 0.18\%  & 62.66 $\pm$ 0.46\%   & 50.12\%  \\ \hline
Q1 & 78.25 $\pm$ 0.22\%  & 48.17 $\pm$ 0.35\%   & 40.27\%  \\ \hline
HT & 86.99 $\pm$ 0.17\%  & 58.95 $\pm$ 0.27\%   & 46.95\%  \\ \hline\hline

W16         &93.36 $\pm$ 0.16 \%  & 72.00 $\pm$ 0.32\% &- \\ \hline
W8         & 93.30 $\pm$ 0.23\%  & 72.71 $\pm$ 0.40\%  &-\\ \hline
W4         & 93.10 $\pm$ 0.11\%  & 72.79 $\pm$ 0.36\%  &-\\ \hline
W2         & 92.80 $\pm$ 0.13\%  & 71.04 $\pm$ 0.22\%  &-\\ \hline
W1          & 91.49 $\pm$ 0.22\%  & 68.27 $\pm$ 0.35\%  &-\\ \hline\hline

A16         & 91.15 $\pm$ 0.31\%  & 65.92 $\pm$ 0.53\%  &-\\ \hline
A8         & 90.86 $\pm$ 0.07\%  & 65.36 $\pm$ 0.34\%  &-\\ \hline
A4         & 90.96 $\pm$ 0.30\%  & 65.67 $\pm$ 0.43\%  &-\\ \hline
A2         & 90.69 $\pm$ 0.11\%  & 64.66 $\pm$ 0.33\%  &-\\ \hline
A1          & 88.70 $\pm$ 0.38\%  & 60.04 $\pm$ 0.48\%  &-\\ \hline
\end{tabular}
\label{tab:quant_base_resnet}
\end{table}}

{\renewcommand{\arraystretch}{1.3}
\begin{table}[bht!]
\centering
\caption{Baseline model accuracies for quantized inputs, weight and activations on CIFAR-10 and CIFAR-100 with VGG11 as baseline}
\begin{tabular}{|c|c|c|}
\hline
Quantization        & CIFAR-10 & CIFAR-100 \\ \hline
FP & 88.58\%  & 55.50\%  \\ \hline
Q8         & 87.74\%  & 56.58\%  \\ \hline
Q6         & 87.61\%  & 55.24\%  \\ \hline
Q4         & 87.10\%  & 54.50\%  \\ \hline
Q2         & 82.58\%  & 50.44\%  \\ \hline
Q1          & 74.10\%  & 37.88\%  \\ \hline
HT       & 80.74\%  & 46.27\%  \\ \hline\hline

W16         & 88.16\%  & 56.58\%  \\ \hline
W8         & 88.16\%  & 55.24\%  \\ \hline
W4         & 87.75\%  & 54.50\%  \\ \hline
W2         & 86.82\%  & 50.44\%  \\ \hline\hline

A16         & 88.45\%  & 57.33\%  \\ \hline
A8         & 88.15\%  & 57.63\%  \\ \hline
A4         & 87.63\%  & 41.12\%  \\ \hline
A2         & 87.00\%  & 57.71\%  \\ \hline
A1         & 74.74\%  & 32.76\%  \\ \hline

\end{tabular}
\label{tab:quant_base_vgg}
\end{table}}

{\renewcommand{\arraystretch}{1.3}
\begin{table}[bht!]
\centering
\caption{Baseline model accuracies for different architectures on CIFAR-10, CIFAR-100 and ImageNet with ResNet18 as baseline }
\begin{tabular}{|c|c|c|c|}
\hline
Arch.       & CIFAR-10 & CIFAR-100 & ImageNet \\ \hline
RN18 & 93.33 $\pm$ 0.13\%  & 72.55 $\pm$ 0.55\%   & 69.76\%  \\ \hline
RN34 & 94.69 $\pm$ 0.07\%  & 75.85 $\pm$ 0.13\%   & 73.31\%  \\ \hline
RN50 & 94.95 $\pm$ 0.13\%  & 77.45 $\pm$ 0.37\%   & -  \\ \hline
RN101& 95.00 $\pm$ 0.14\%  & 76.98 $\pm$ 0.28\%   & 77.37\%  \\ \hline
VGG11& 88.58 $\pm$ 0.21\%  & 58.75 $\pm$ 0.62\%   &69.02\%  \\ \hline
VGG19& 89.61 $\pm$ 0.29\%  & 61.72 $\pm$ 0.75\%   &72.38\%  \\ \hline
VGG11BN& 89.94 $\pm$ 0.18\%  & 63.68 $\pm$ 0.25\%   &-  \\ \hline
VGG19BN& 91.34 $\pm$ 0.17\%  & 66.04 $\pm$ 0.56\%   &-  \\ \hline
DN121& -  & -   & 74.43\%  \\ \hline
WRN50\_2& -  & -   & 78.47\%  \\ \hline
\end{tabular}
\label{tab:arch_base}
\end{table}}

\section{Architecture Analysis} \label{adx:Arch}
\begin{figure*}[!htb]
\centering
    \begin{subfigure}[b]{0.3\textwidth}
        \centering
        \includegraphics[scale=0.3]{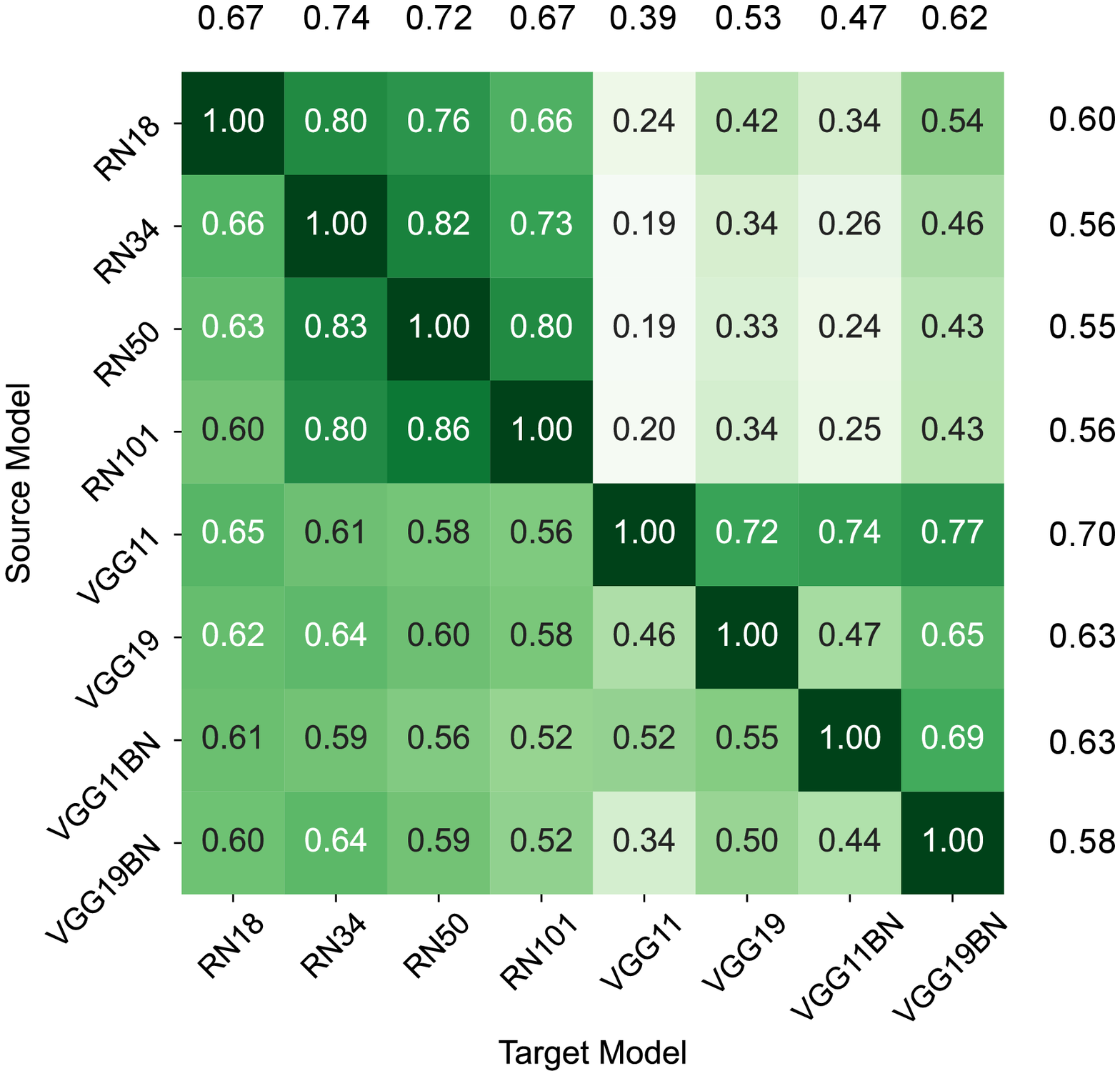}
        \caption{Architecture analysis, under PGD attack, 40 iterations}
        \label{fig:cifar100arch}
    \end{subfigure}
    \hfill
        \begin{subfigure}[b]{0.3\textwidth}
        \centering
        \includegraphics[scale=0.3]{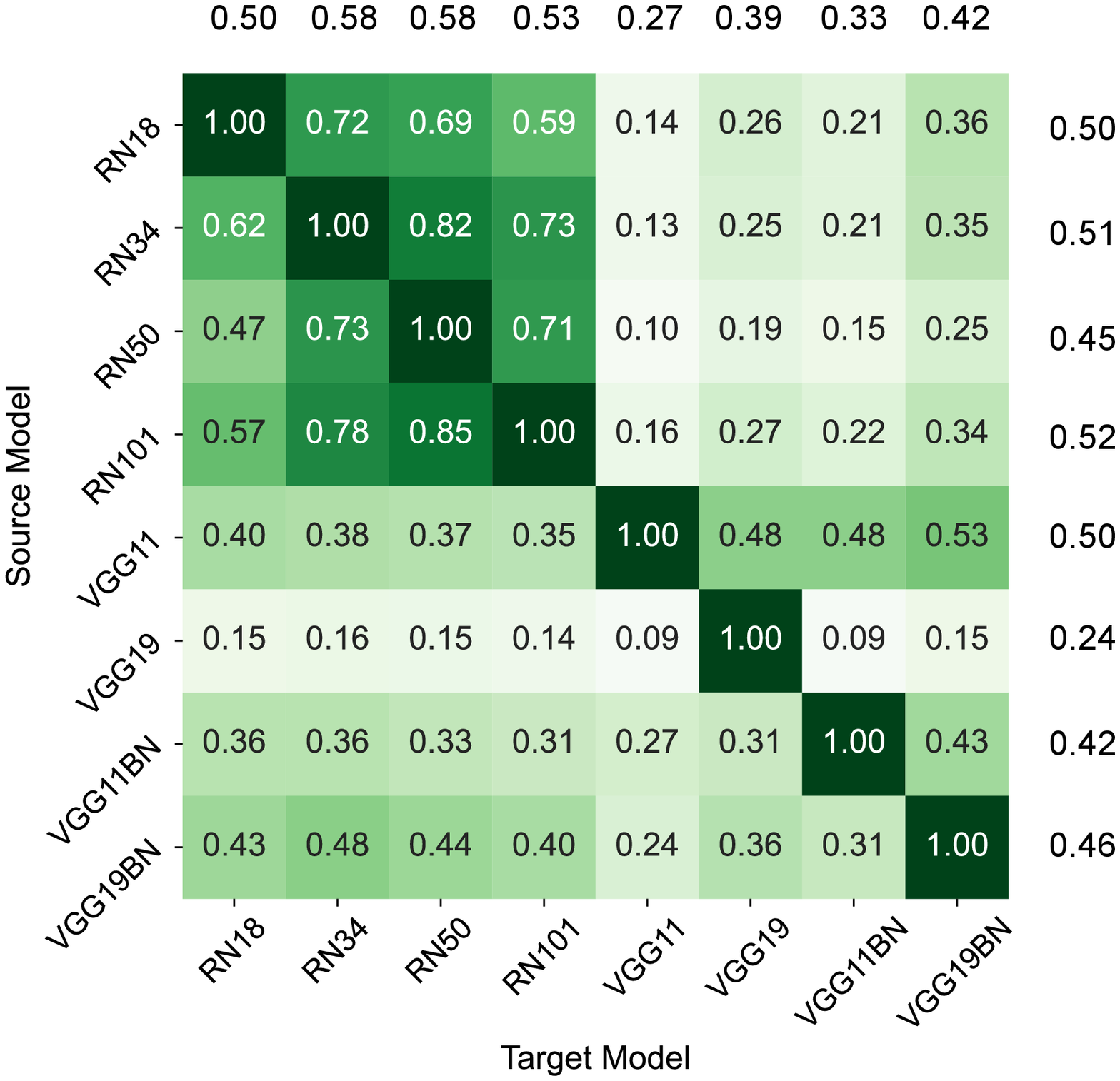}
        \caption{Architecture analysis, under Carlini Wagner $L_{2}$ ($\kappa$=30) attack}
        \label{fig:cifar100arch_cw}
    \end{subfigure}
    \hfill
    \begin{subfigure}[b]{0.3\textwidth}
        \centering
        \includegraphics[scale=0.31]{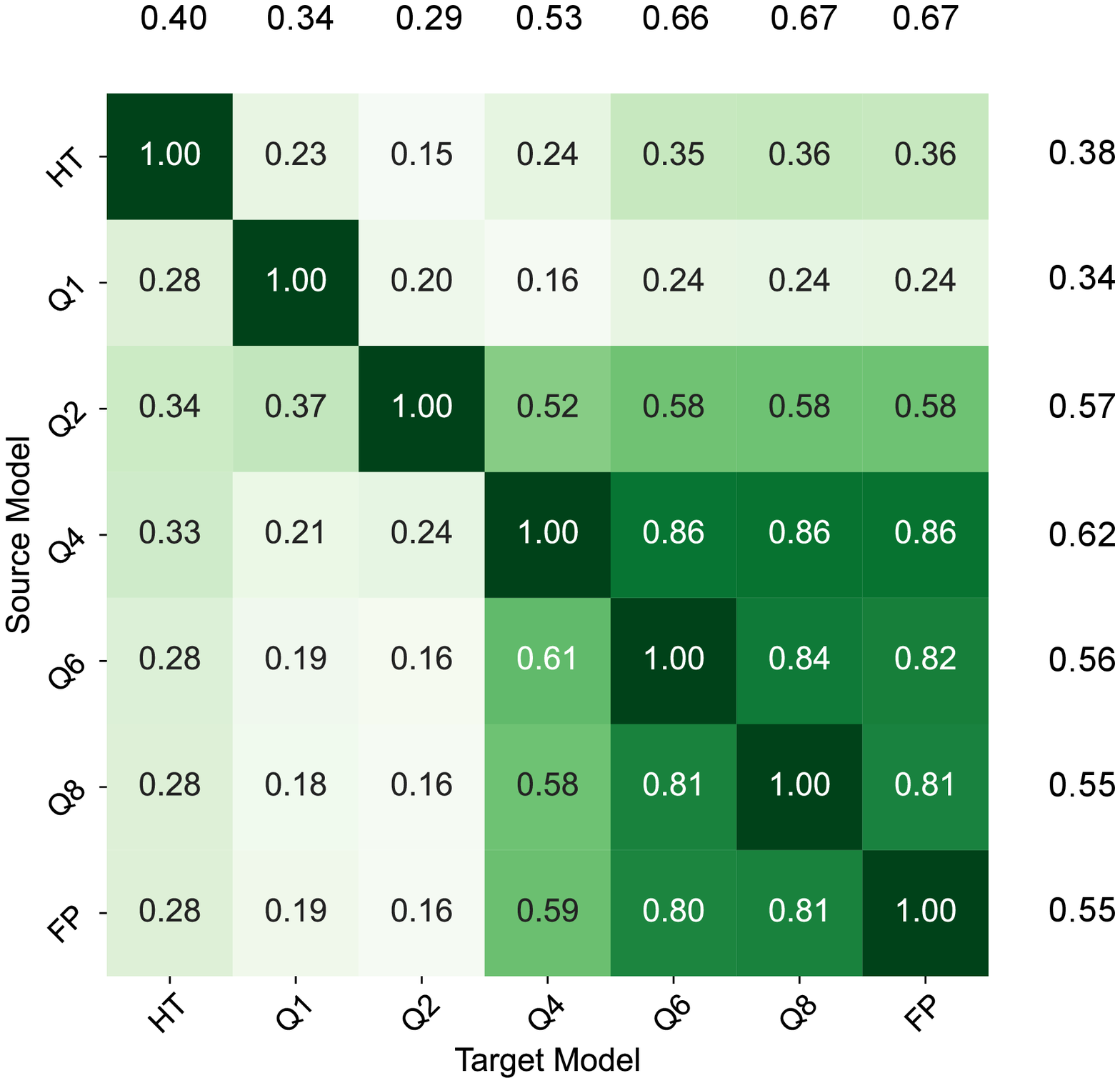}
        \caption{Input quantization analysis under PGD attack, ResNet18 base model.}
        \label{fig:cifar100Quant}
    \end{subfigure}
    \caption{Number of adversarial images transferred from source to target on CIFAR-100 dataset under PGD attack averaged over 5 differently seeded models for each architecture (a) and input quantization (b).}
\end{figure*}
\begin{figure}[ht]{}
    \centering
    \includegraphics[scale=0.34]{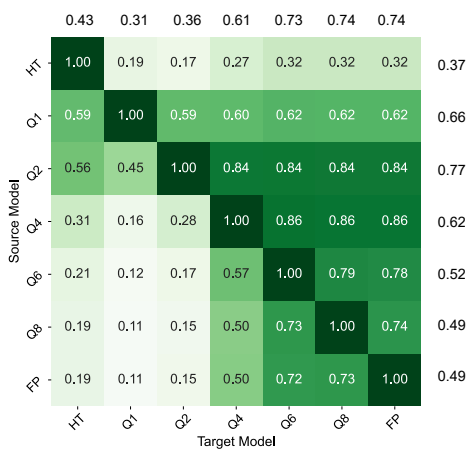}
    \caption{Input quantization analysis under Carlini Wagner $L_{2}$ attack ($\kappa$=30), ResNet18 base model.}
    \label{fig:cifar100Quant_CW}
\end{figure}
Figure \ref{fig:cifar100arch} shows the number of adversarial images transferred  from source to target for various architectures on CIFAR-100. Figure \ref{fig:cifar100arch} can be interpreted by analyzing the 4 quadrants, with each quadrant representing a family of source or target model architectures (ResNet or VGG variants). The top right quadrant of Figure \ref{fig:cifar100arch} is lighter than the bottom left quadrant. This implies that adversarial images generated on VGG are more transferable to ResNets than the other way around.

\begin{figure}[!htb]
%   \begin{minipage}{0.48\textwidth}
      \centering
      \includegraphics[scale=0.3]{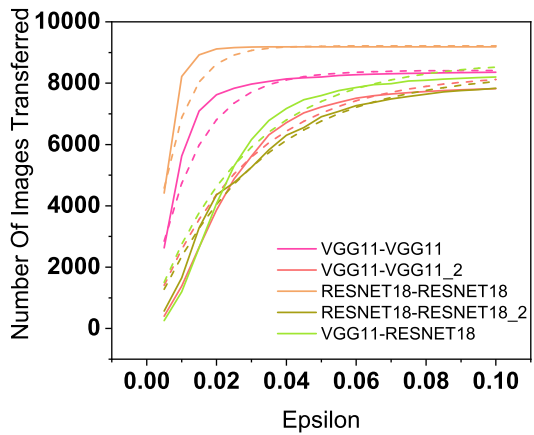}
      \caption{Predicted (dotted) and actual (solid) number of images transferred vs attack strength $\epsilon$ for CIFAR-10 from (source-target). Here \_2 refers to differently seeded model.}
        \label{fig:cap_charging_curve}
%   \end{minipage}\hfill
%   \begin{minipage}{0.48\textwidth}
%       \centering
%       \includegraphics[scale=0.45]{images/TM.eps}
%       \caption{Transferability metric for various models on CIFAR-10 at $\epsilon$ of 8/255, obtained from equation \ref{eq:TM}, fit using 4 datapoints. The empirical counterpart is found in Fig \ref{fig:cifar100Quant}}
%       \label{fig:tm_cifar10}
%   \end{minipage}
 \end{figure}

\begin{figure*}[!htb]
    \centering
    \begin{subfigure}[b]{0.48\textwidth}
    \centering
        \includegraphics[scale=0.30]{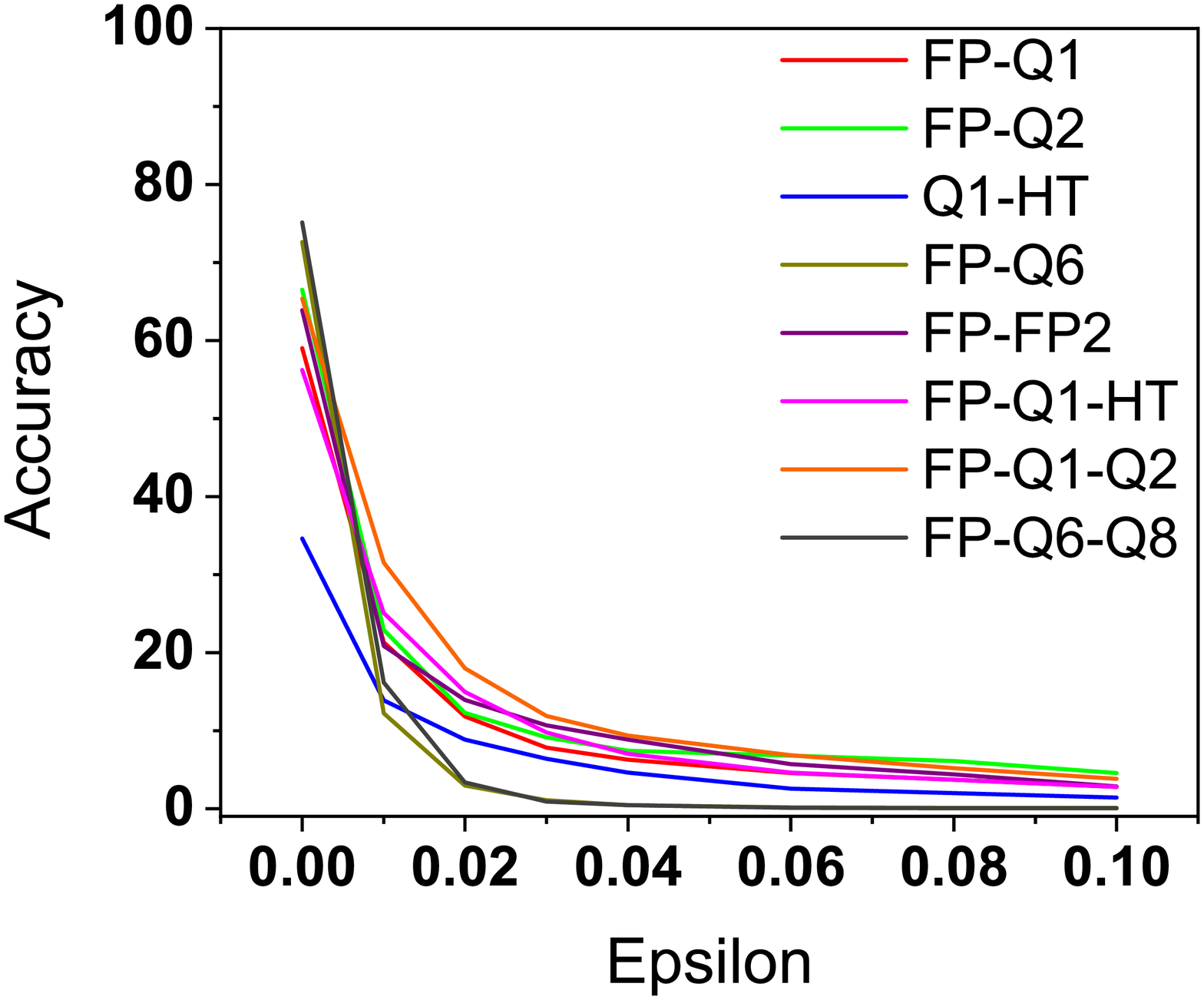}
        \caption{CIFAR-100 Gradient Average}
        \label{fig:cifar100avg}
    \end{subfigure}
    \hfill
    \begin{subfigure}[b]{0.48\textwidth}
    \centering
        \includegraphics[scale=0.3]{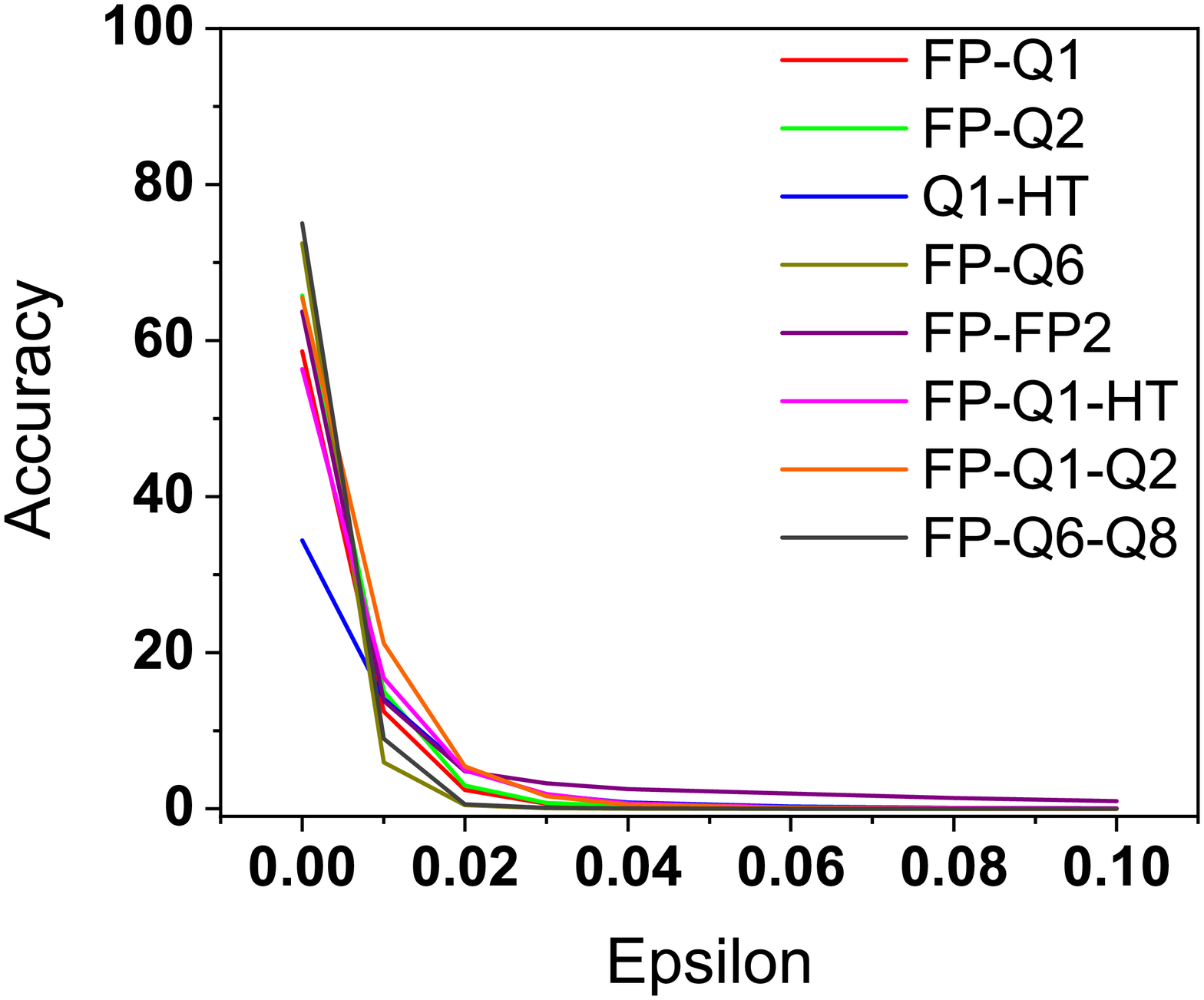}
        \caption{CIFAR-100 Unanimous Gradient Direction}
        \label{fig:cifar100signall}
    \end{subfigure}
\caption{Results for ensembles under different attacks for CIFAR-100. }
\label{fig:EnsembleAccuracy}
\end{figure*}

\section{Input Quantization} \label{adx:input_quant}

\begin{figure*}[!htb]
\centering
    \begin{subfigure}[b]{0.3\textwidth}
        \centering
        \includegraphics[scale=0.3]{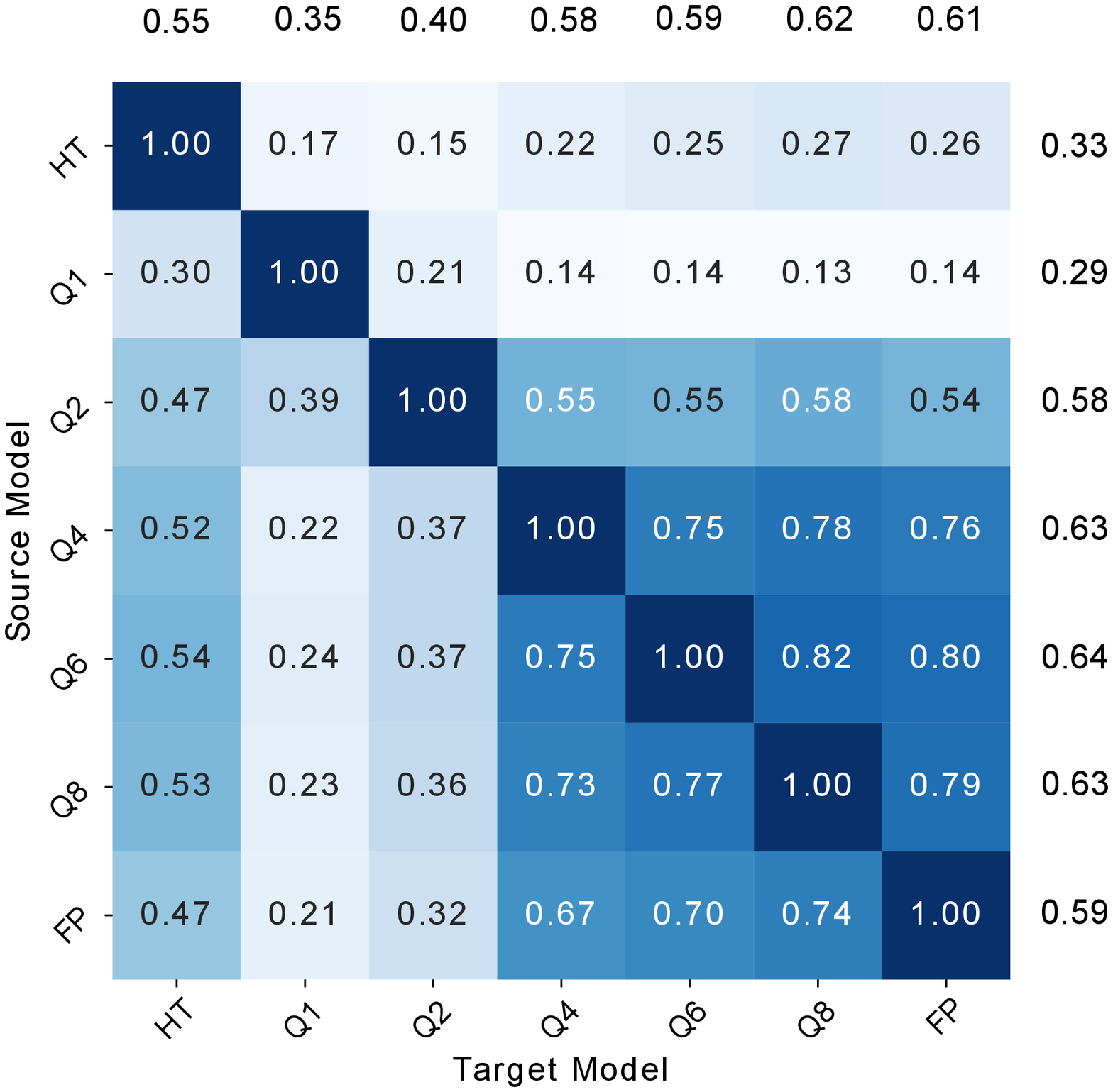}
        \caption{Input quantization analysis, VGG11 base model, CIFAR-10}
        \label{fig:cifar10QuantVgg}
        \vspace{3mm}
    \end{subfigure}
    \hfill
    \begin{subfigure}[b]{0.3\textwidth}
        \centering
        \includegraphics[scale=0.3]{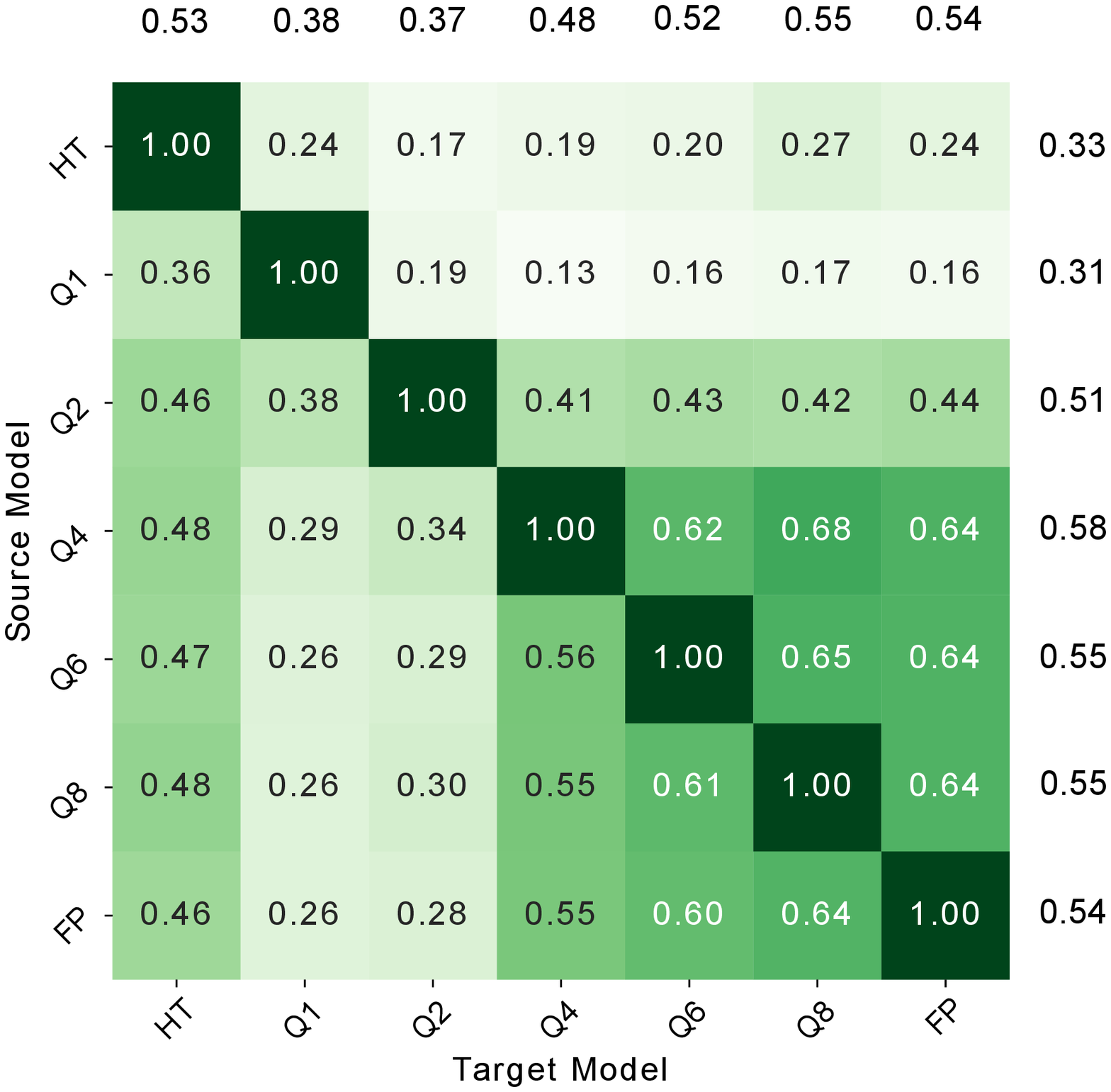}
        \caption{Input quantization analysis, VGG11 base model, CIFAR-100}
        \label{fig:cifar100QuantVgg}
        \vspace{3mm}
    \end{subfigure}
    \hfill
    \begin{subfigure}[b]{0.3\textwidth}
        \centering
        \includegraphics[scale=0.31]{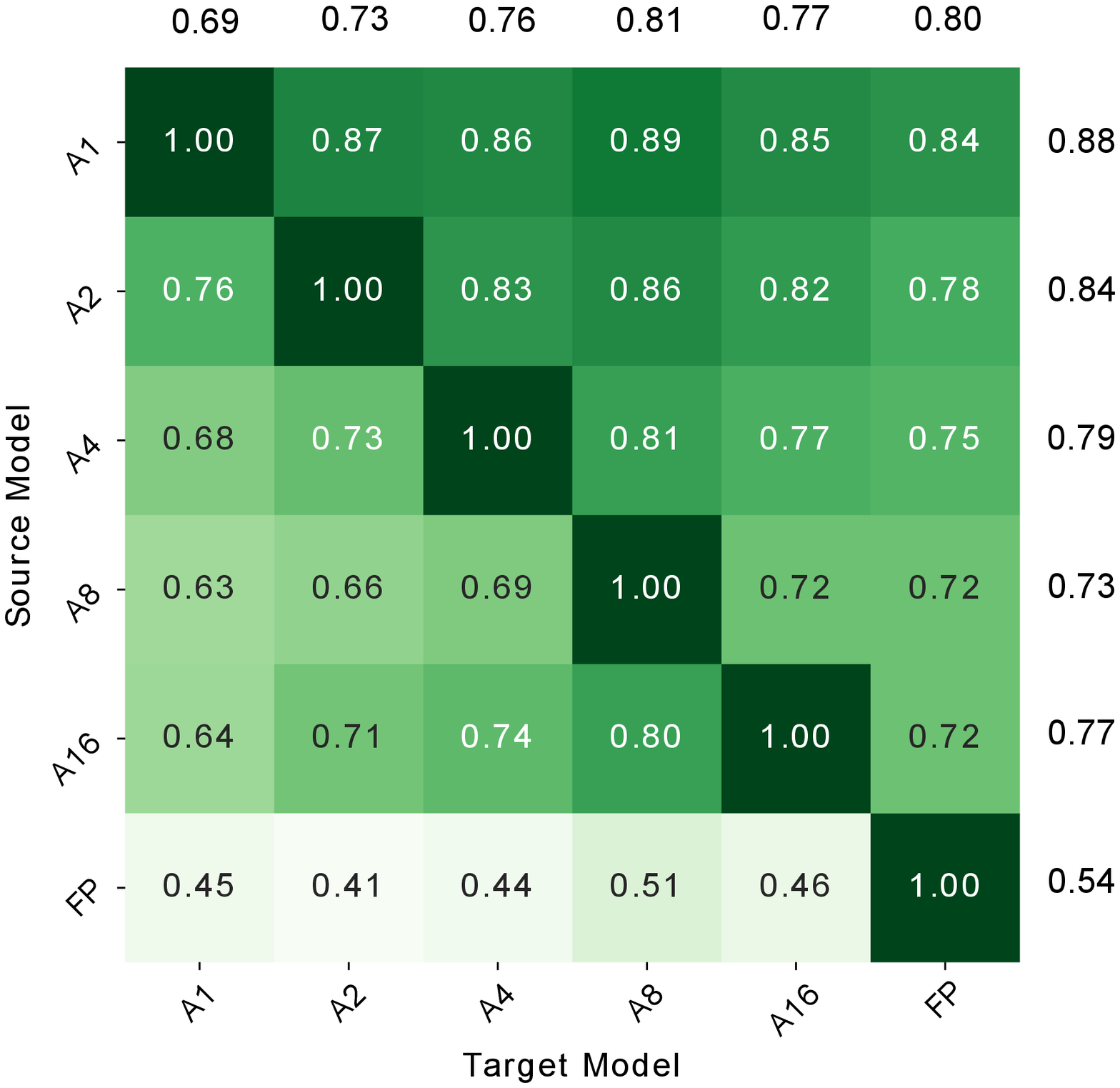}
        \caption{Activation quantized ResNet18 base model, CIFAR-100}
        \label{fig:cifar100Act}
        \vspace{3mm}
    \end{subfigure} 
    \hfill
    \begin{subfigure}[b]{0.3\textwidth}
    \centering
        \includegraphics[scale=0.3]{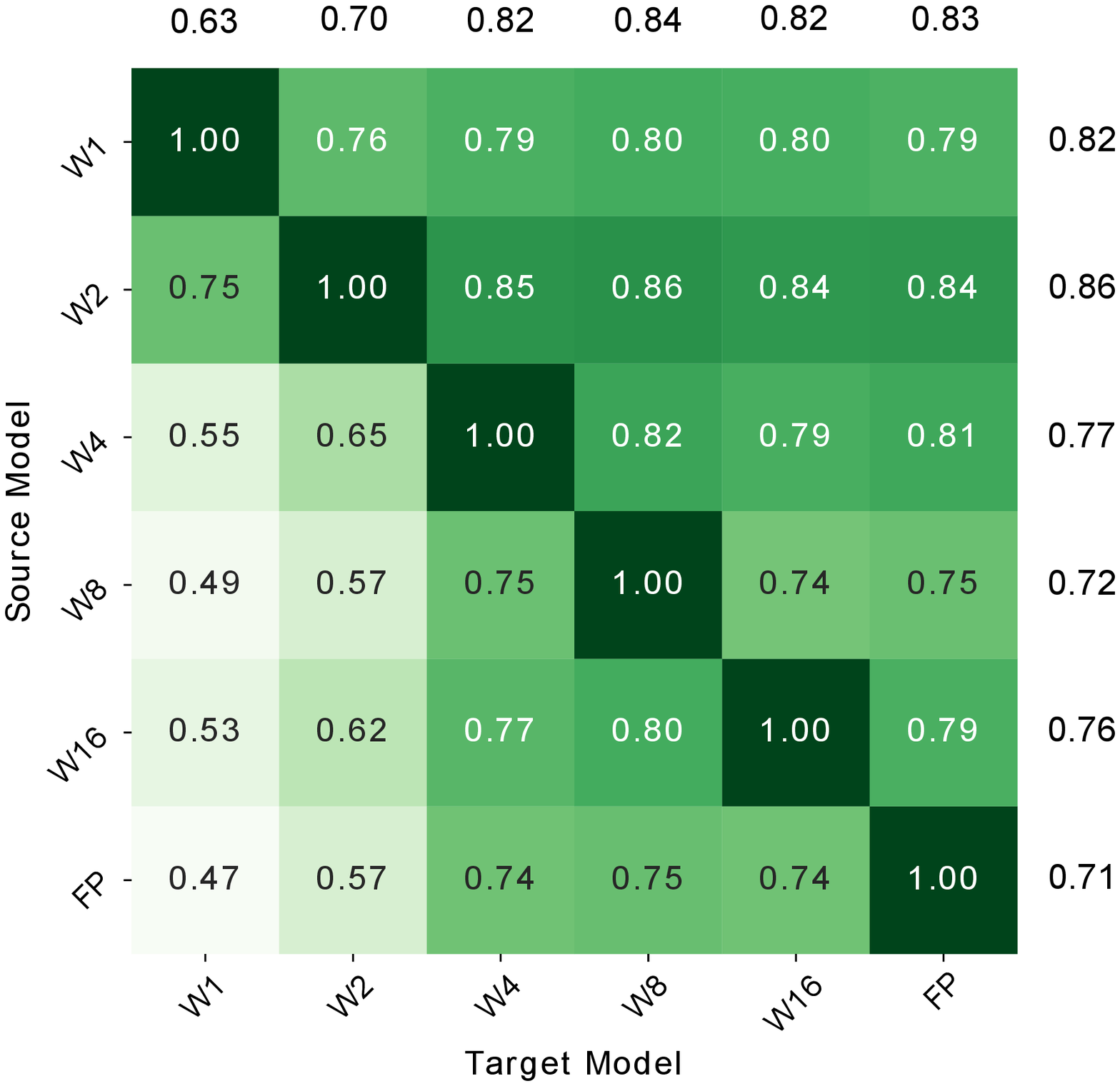}
        \caption{Weight quantized ResNet18 base model, CIFAR-100}
        \label{fig:cifar100Weight}
    \end{subfigure}
    \hfill
    \begin{subfigure}[b]{0.3\textwidth}
        \centering
        \includegraphics[scale=0.29]{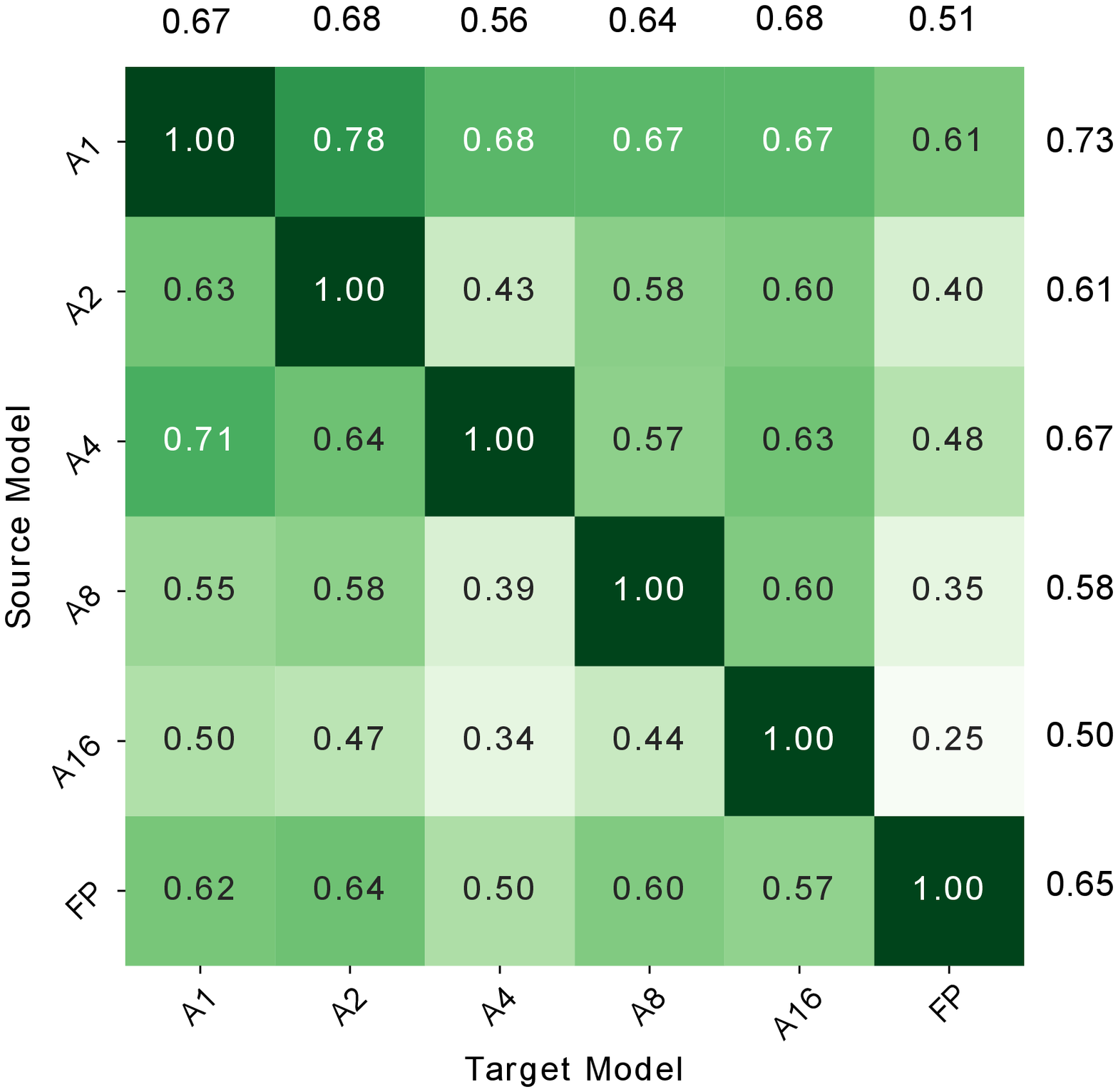}
        \caption{Activation quantized VGG11 base model, CIFAR-100\footnotemark[1]}
        \label{fig:cifar100ActVGG}
    \end{subfigure} 
    \hfill
    \begin{subfigure}[b]{0.3\textwidth}
    \centering
        \includegraphics[scale=0.3]{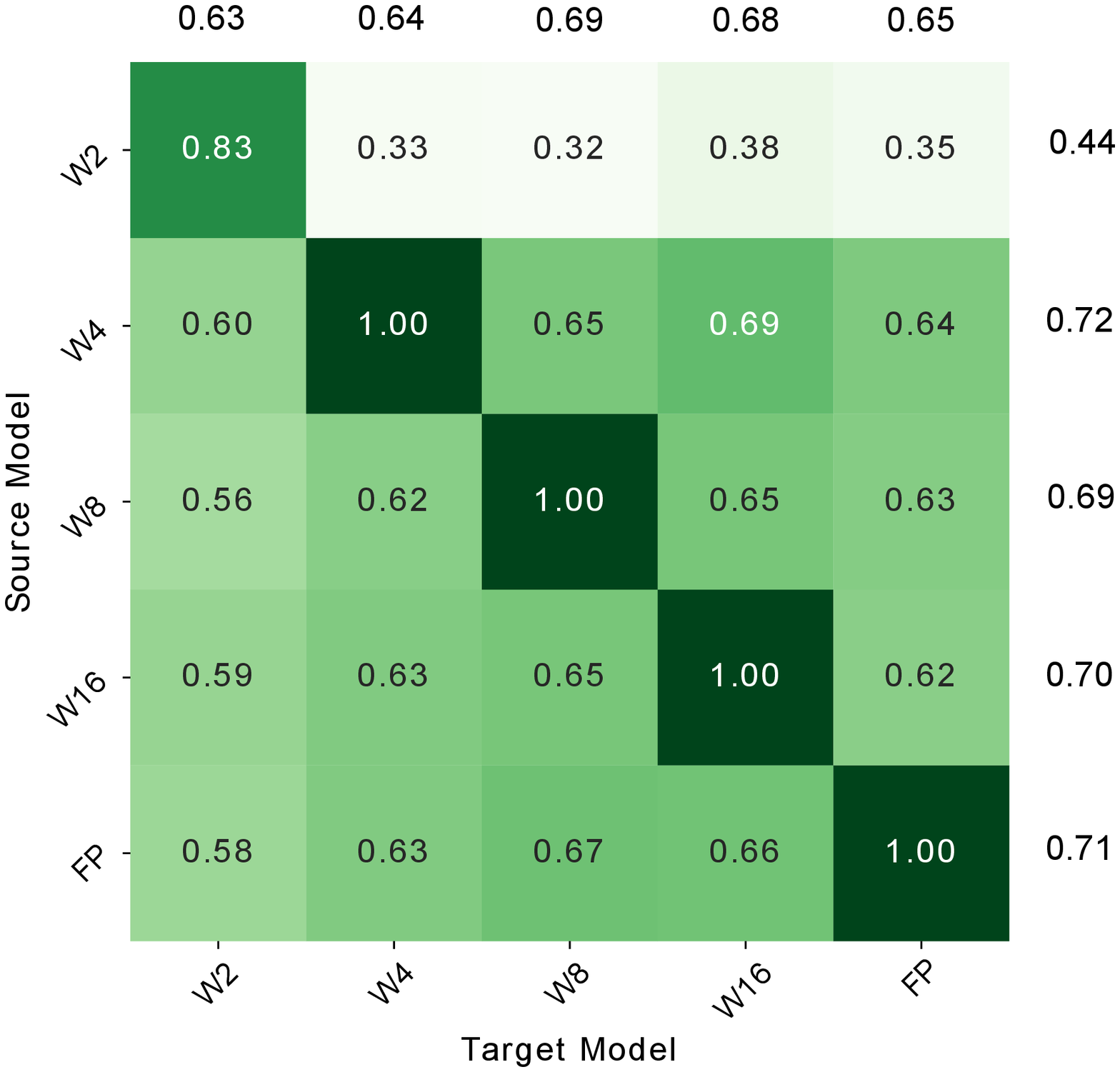}
        \caption{Weight quantized VGG11 models\footnotemark[1]}
        \label{fig:cifar100WeightVGG}
    \end{subfigure}
\caption{Number of adversarial images transferred from source to target on CIFAR-10 and CIFAR-100 dataset}
\label{fig:weight_act_quant_cifar100}
\end{figure*}

\footnotetext[1]{The quantization results presented on VGG11 are from single runs, unlike other results presented which are averaged over multiple seeds.}

 Figure \ref{fig:cifar100Quant}, Figure \ref{fig:cifar100Quant_CW}, Figure \ref{fig:cifar10QuantVgg} and Figure \ref{fig:cifar100QuantVgg} show the variation in transferability due input quantization on CIFAR-10 and CIFAR-100 with ResNet and VGG as the base models.

\section{Weight and Activation Quantization} \label{adx:weight_act_quant}
Figure \ref{fig:weight_act_quant_cifar100} shows the confusion matrices for activation and weight quantized models on CIFAR-100. 
%Figure \ref{fig:weight_act_quant_cifar10} shows the confusion matrices for activation and weight quantized VGG11 models on CIFAR-10.
Figure \ref{fig:cifar100Act} and Figure \ref{fig:cifar100Weight} use ResNet18 as their base models while Figure \ref{fig:cifar100ActVGG} and Figure \ref{fig:cifar100WeightVGG} use VGG11 as their base model.

% \begin{figure*}[!htb]
%     \centering
%     \begin{subfigure}[b]{0.48\textwidth}
%         \centering
%         \includegraphics[scale=0.3]{images/cifar10ActVggTrans.eps}
%         \caption{Activation quantized VGG11 models\footnotemark[1]}
%         \label{fig:cifar10ActVgg}
%     \end{subfigure} 
%     \hfill
%     \begin{subfigure}[b]{0.48\textwidth}
%     \centering
%         \includegraphics[scale=0.3]{images/cifar10WeightVggTrans.eps}
%         \caption{Weight quantized VGG11 models\footnotemark[1]}
%         \label{fig:cifar10WeightVgg}
%     \end{subfigure}
% \caption{Number of adversarial images transferred from source to target on CIFAR-10 dataset}
% \label{fig:weight_act_quant_cifar10}
% \end{figure*}

\section{Transferability Metric and Attack Strength} \label{adx:transfer_metric}
In this section, we analyze the variation of transferability with respect to the attack strength $\epsilon$. We plot the number of images that transfer between different source-target models, referred to as $f_{st}(\epsilon)$, for 20 different values of $\epsilon$ in Fig. \ref{fig:cap_charging_curve}. We observe that these curves cross one another and therefore the trends of transferability across models cannot be generalized from the observations at a single $\epsilon$. However, we are interested in plotting this curve with the fewest possible measurements, such that it would have good predictive power over a range of $\epsilon$ values. To do this, we note that each plot visually resembles the CDF of an exponential distribution. The same is true for the number of images generated on a model, denoted by $f_{ss}$. Hence, we characterize both functions in the following form:
\begin{align}
f_{st}(\epsilon)&=a(1-e^{b\epsilon}) \\
f_{ss}(\epsilon)&=a'(1-e^{b'\epsilon})
\end{align}
where $a$, $a'$, $b$ and $b'$ are the parameters for fitting the data, obtained experimentally. We find the parameters of the equation using a few datapoints. Figure \ref{fig:cap_charging_curve} shows the empirical curve obtained using 20 points, and the predicted curve, fit using 4 points.  We observe that they align well and we get $<$ 5\% Root Mean Square Error (RMSE) for the fit.

Armed with these observations, model transferability as a function of attack strength $\epsilon$. Equation \ref{eq:TM} captures transferability as a ratio of the number of images that transfer to the target to the number of images that were generated at the source. Equivalently our model of transferability normalizes $f_{st}$ by $f_{ss}$ and gives us the transferability metric, TM:
\begin{equation}
TM(\epsilon)= \frac{f_{st}(\epsilon)}{f_{ss}(\epsilon)}
\label{eq:TM_eps}
\end{equation}
The transferability metric is a number between [0, 1] and represents a quantitative measure of the transferability between a given pair of models. The constants in both equations capture the effects of dataset, architecture, input quantization etc. and can be estimated with a few datapoints. For example, we use 4 datapoints to estimate the constants for various quantizations for CIFAR-10. 

%The predicted values are shown in Figure \ref{fig:tm_cifar10} and the experimentally obtained values are shown in Figure \ref{fig:cifar10Quant}. The close match between Figure \ref{fig:tm_cifar10} and \ref{fig:cifar10Quant} shows the effectiveness of the proposed transferability metric in predicting adversarial input transferability. 
Table \ref{tab:TMConf} shows the standard error (RMSE) for the fit used to calculate the transferability metric. The average difference between the predicted and actual values is 0.055 or 5.5\% for CIFAR-10 dataset on ResNet18 base model.

{\renewcommand{\arraystretch}{1.5}
\begin{table}[t]
\centering
\caption{Transferability Metric Fit Confidence for CIFAR10} 
\begin{adjustbox}{width=0.93\columnwidth}
\begin{tabular}{|c|c|c|c|}
\hline
\multirow{2}{*}{Source Model} &  \multirow{2}{*}{Target Model} & Standard Error &Standard Error  \\ 
&&Train (RMSE)&Test (RMSE)\\ \hline
VGG11& VGG11 & 114.41& 505.23      \\ \hline
ResNet18& ResNet18 &38.67 & 480.64\\ \hline
VGG11& ResNet18 & 661.34& 516.65          \\ \hline
ResNet18 &VGG11&  44.58& 46.72      \\ \hline
VGG11 Seed 1& VGG11 Seed 2 & 529.92& 405.56 \\ \hline
ResNet18 Seed 1& ResNet18 Seed 2 & 358.37& 248.17 \\ \hline
FP & Q1 & 12.63&47.11\\ \hline
Q1 & FP & 352.34& 423.61\\ \hline
\end{tabular}
\end{adjustbox}
\label{tab:TMConf}
\end{table}}

\section{Results for differrent Ensembles} \label{adx:Results_Ensemble}
% From Figure \ref{fig:difference} and \ref{fig:imagenetError} 
We observe that Q1, Q2 and HT are robust models for CIFAR-10, CIFAR-100 and ImageNet datasets.
The result for ensemble on CIFAR-10 and CIFAR-100 under different attacks are shown in Figure \ref{fig:EnsembleAccuracy}.

\end{document}